\documentclass[10pt,twocolumn,letterpaper]{article}

\usepackage[pagenumbers]{cvpr} 

\usepackage{graphicx}
\usepackage{amsmath}
\usepackage{amssymb}
\usepackage{booktabs}

\usepackage[pagebackref,breaklinks,colorlinks]{hyperref}

\usepackage{times}
\usepackage{epsfig}
\usepackage{graphicx}
\usepackage{amsmath}
\usepackage{amssymb}
\usepackage{multirow}
\usepackage{microtype}
\usepackage[normalem]{ulem}
\useunder{\uline}{\ul}{}
\usepackage{color}
\usepackage{nicefrac}
\usepackage{comment}
\usepackage{enumitem}
\usepackage{adjustbox}
\usepackage{mathabx}
\usepackage{tabu}
\usepackage{diagbox}
\usepackage{cuted}
\usepackage{capt-of} 
\usepackage[table,x11names]{xcolor}
\usepackage{tabularx}
\usepackage{tablefootnote}

\usepackage{amssymb}
\usepackage{lipsum}

\usepackage{pifont}
\usepackage{algorithm}
\usepackage{algpseudocode} 
\usepackage[T1]{fontenc}
\usepackage{array}
\newcolumntype{x}[1]{>{\centering\arraybackslash\hspace{0pt}}p{#1}}
\newcolumntype{u}[1]{>{\raggedleft\arraybackslash\hspace{0pt}}p{#1}}

\usepackage[capitalize]{cleveref}
\crefname{section}{Sec.}{Secs.}
\Crefname{section}{Section}{Sections}
\Crefname{table}{Table}{Tables}
\crefname{table}{Tab.}{Tabs.}

\newcommand{\citep}{\cite}%
\newcommand{\citet}{\cite}%

\usepackage{overpic}
\usepackage{enumitem} 
\usepackage{overpic} 
\usepackage{color}

\definecolor{turquoise}{cmyk}{0.65,0,0.1,0.3}
\definecolor{purple}{rgb}{0.65,0,0.65}
\definecolor{dark_green}{rgb}{0, 0.5, 0}
\definecolor{orange}{rgb}{0.8, 0.6, 0.2}
\definecolor{red}{rgb}{0.8, 0.2, 0.2}
\definecolor{darkred}{rgb}{0.6, 0.1, 0.05}
\definecolor{blueish}{rgb}{0.0, 0.3, .6}
\definecolor{light_gray}{rgb}{0.7, 0.7, .7}
\definecolor{pink}{rgb}{1, 0, 1}
\definecolor{greyblue}{rgb}{0.25, 0.25, 1}

\usepackage{blindtext}

\renewcommand{\paragraph}[1]{\vspace{1em}\noindent\textbf{#1}.}
\begin{document}
\title{LANA: Latency Aware Network Acceleration}

\author{TBD\\
NVIDIA\\
{\tt\small dannyy@nvidia.com}
}




\author{
Pavlo Molchanov\\
NVIDIA\\
\and
Jimmy Hall\\
Microsoft Research\\
\and
Hongxu Yin\\
NVIDIA\\
\and 
Jan Kautz\\
NVIDIA\\  
\and
\quad  \\
\quad  \\
\and
Nicolo Fusi\\
Microsoft Research\\
\and
Arash Vahdat\\
NVIDIA
}

\maketitle

\begin{abstract}
We introduce latency-aware network acceleration (LANA) – an approach that builds on neural architecture search techniques and teacher-student distillation to accelerate neural networks. LANA consists of two phases: in the first phase, it trains many alternative operations for every layer of the teacher network using layer-wise feature map distillation. In the second phase, it solves the combinatorial selection of efficient operations using a novel constrained integer linear optimization (ILP) approach. ILP brings unique properties as it (i) performs NAS within a few seconds to minutes, (ii) easily satisfies budget constraints, (iii) works on the layer-granularity, (iv) supports a huge search space $O(10^{100})$, surpassing prior search approaches in efficacy and efficiency.

In extensive experiments, we show that LANA yields efficient and accurate models constrained by a target latency budget, while being significantly faster than other techniques. We analyze three popular network architectures: EfficientNetV1, EfficientNetV2 and ResNeST, and achieve up to $3.0\%$ accuracy improvement for all models when compressing larger models to the latency level of smaller models. LANA achieves significant speed-ups (up to 5$\times$) with minor to no accuracy drop on GPU and CPU. The code will be available soon.
\end{abstract}

\vspace{-0.5cm}
\section{Introduction}
\label{sec:intro}
In many applications, we may have access to a neural network that satisfies desired performance needs in terms of accuracy but is computationally too expensive to deploy. The goal of hardware-aware network acceleration~\cite{rastegari2016xnor, zhang1707shufflenet, sze2017efficient, he2018amc, cai2020once, chamnet} is to accelerate a given neural network such that it meets efficiency criteria on a device without sacrificing accuracy dramatically. Network acceleration plays a key role in reducing the operational cost, power usage, and environmental impact of deploying deep neural networks in real-world applications.

The current network acceleration techniques can be grouped into: \textit{(i) pruning} that removes inactive neurons~\citep{mozer1989skeletonization, lecun1990optimal, hanson1989comparing, chauvin1989back, molchanov2016pruning, molchanov2019importance, thinet, liu2018rethinking, nisp, he2018soft, he2019filter, frankle2018lottery, he2017channel, zhu2016trained, lebedev2016fast,li2016pruning, ICLR2018, he2018adc, louizos2017learning, neklyudov2017structured, he2016identity,huang2017densely, gordon2018morphnet, yu2017nisp, he2018progressive, molchanov2017variational, blalock2020state}, (ii) \textit{compile-time optimization}~\cite{Ryoo2008OptimizationPA} \textit{or kernel fusion}~\citep{Wang2010KernelFA, ding2021repvgg, ding2019acnet, zagoruyko2017diracnets} that combines multiple operations into an equivalent operation, (iii) \textit{quantization} that reduces the precision in which the network operates at~\citep{wang2019haq, dong2019hawq, shen2020qbert, cai2020zeroq, Choi2018PACTPC, Park2018ValueawareQF, Wu2018MixedPQ, Zhang2018LQNetsLQ, Krishnamoorthi2018QuantizingDC}, and \textit{(iv) knowledge distillation} that distills knowledge from a larger \textit{teacher} network into a smaller \textit{student} network~\citep{hinton2015distilling, sau2016deep, xu2017training,yin2020dreaming,mishra2017apprentice, nayak2019zero, belagiannis2018adversarial}. The approaches within (i) to (iii) are restricted to the underlying network operations and they do not change the architecture. Knowledge distillation changes the network architecture from teacher to student, however, the student design requires domain knowledge and is done usually manually.

\begin{table*}[t]
\scriptsize
\centering
\caption{Related method comparison. Time is mentioned in GPU hours by \texttt{h}, or ImageNet epochs by \texttt{e}. Our method assumes 197 candidate operations for the full pool, and only 2 (teacher and identity) for zero shot mode. \(L\) is the number of target architectures.} 
\label{tab:related_work}\vskip -5pt
\resizebox{0.9\textwidth}{!}{
\begin{tabular}{l||c|c|c||c|c|c||c}
\toprule
Method & {Knowledge} & Diverse & Design Space & Pretrain & Search & Train & Total  \\
 & Distillation & Operators & Size & Cost & Cost & Cost & Cost  \\
\hline
&  &   & & \multicolumn{4}{c}{To Train \(L\) Architectures}\\
\hline
Once-For-All \cite{cai2020once} & None & & \(> O\left(10^{19}\right)\) & 1205\texttt{e} & 40\texttt{h}& 75\texttt{e}$L$ & 1205\texttt{e} + 75\texttt{e}$L$\\ 
\hline
AKD \cite{Liu_2020_CVPR} & Network & \checkmark & \(> O\left(10^{13}\right)\) &
0 & 50000\texttt{e}\(L\) & 400\texttt{e}\(L\) & 50400\texttt{e}\(L\) \\ 
DNA \cite{Li_2020_CVPR} & Block &  & \(> O \left(10^{15}\right)\) & 320\texttt{e} & 14\texttt{h}$L$ & 450\texttt{e}$L$ & 320\texttt{e} + 450\texttt{e}$L$\\
DONNA \cite{moons2020distilling} & Block & \checkmark & \(> O\left(10^{13}\right)\) & 1920\texttt{e}\tablefootnote{can potentially be improved by parallelization} & 1500\texttt{e} + <1\texttt{h}$L$ & 50\texttt{e}$L$ & 3420\texttt{e} + 50\texttt{e}$L$ \\
\hline
This Work & Layer & \checkmark & \(>O\left(10^{100}\right)\) & 197\texttt{e} & $<$1\texttt{h}$L$ & 100\texttt{e}$L$ &  197\texttt{e} + 100\texttt{e}$L$\\
This Work - Zero shot& None &  & \(>O\left(10^{2}\right)\) & 0 & $\sim$0 & 100\texttt{e}$L$ & 100\texttt{e}$L$\\
\bottomrule
\end{tabular}
}
\end{table*}

In this paper, we propose latency-aware network transformation (LANA), a network acceleration framework that automatically replaces inefficient operations in a given network with more efficient counterparts. Given a convolutional teacher network, we formulate the problem as searching in a large pool of candidate operations to find efficient operations for different layers of the teacher. The search problem is combinatorial in nature with a space that grows exponentially with the depth of the network. To solve this problem, we can turn to neural architecture search (NAS)~\citep{zoph2016neural, zoph2018learning, tan2018mnasnet, cai2018proxylessnas, pham2018efficient, vahdat2020unas}, which has been proven successful in discovering novel architectures. However, existing NAS solutions are computationally expensive, and usually handle only a small number of candidate operations (ranging from 5 to 15) in each layer and they often struggle with larger candidate pools.

To tackle the search problem with a large number of candidate operations in an efficient and scalable way, we propose a two-phase approach. In the first phase, we define a large candidate pool of operations ranging from classic residual blocks \citep{he2016deep} to recent blocks~\citet{dosovitskiy2020image, ding2021repvgg, srinivas2021bottleneck, bello2021lambdanetworks}, with varying hyperparameters. 
Candidate operations are pretrained to mimic the teacher's operations via a simple layer-wise optimization.
Distillation-based pretraining enables a very quick preparation of all candidate operations, offering a much more competitive starting point for subsequent searching.

In the second phase, we search among the pre-trained operations as well as the teacher's own operations to construct an efficient network. Since our operation selection problem can be considered as searching in the proximity of the teacher network in the architecture space, we assume that the accuracy of a candidate architecture can be approximated by the teacher's accuracy and a simple linear function that measures changes in the accuracy for individual operations. Our approximation allows us to relax the search problem into a constrained integer linear optimization problem that is solved in a few seconds. As we show extensively in our experiments, such relaxation can drastically cut down on the cost of our search and it can be easily applied to a huge pool of operations (197 operations per layer), while offering improvements in model acceleration by a large margin. 

In summary, we make the following contributions: \textbf{(i)} We propose a simple two-phase approach for accelerating a teacher network using NAS-like search. \textbf{(ii)} We propose an effective search algorithm using constrained integer optimization that can find an architecture in seconds tailored to our setting where a fitness measure is available for each operation. \textbf{(iii)} We examine a large pool of operations including the recent vision transformers and new variants of convolutional networks. We provide insights into the operations selected by our framework and into final model architectures.

\vspace{-0.0cm}
\subsection{Related Work}

Since our goal is to discover new efficient architectures, in this section we focus on related NAS-based approaches.  

\textbf{Hardware-aware NAS:} The goal of hardware-aware NAS is to 
design efficient and accurate architectures from scratch while targeting a specific hardware platform. This has been the focus of an increasingly large body of work on multiobjective neural architecture search \citep{cai2018proxylessnas, Tan_2019_CVPR, Wu_2019_CVPR, yang2018netadapt, veniat2017learning, yang2018synetgy, wu2018mixed, vahdat2020unas}. The goal here is to solve an optimization problem maximizing accuracy while meeting performance constraints specified in terms of latency, memory consumption or number of parameters. Given that the optimization problem is set up from scratch for each target hardware platform, these approaches generally require the search to start from scratch for every new deployment target (\emph{e.g.}, GPU/CPU family) or objective, incurring a search cost that increases linearly as the number of constraints and targets increases. \citet{cai2020once} circumvents this issue by training a supernetwork containing every possible architecture in the search space, and then applying a progressive shrinking algorithm to produce multiple high-performing architectures. This approach incurs a high pretraining cost, but once training is complete, new architectures are relatively inexpensive to find. On the other hand, the high computational complexity of pretraining limits the number of operations that can be considered. Adding new operations is also costly, since the supernetwork must be pretrained from scratch every time a new operation is added. 

\textbf{Teacher-based NAS:} Our work is more related to the line of work that focuses on modifying \emph{existing} architectures. Approaches in this area build on teacher-student knowledge distillation, performing multiobjective NAS on the student to mimic the teacher network. Pioneering works \cite{Liu_2020_CVPR}, \cite{Li_2020_CVPR} and \cite{moons2020distilling} demonstrated great benefits of teacher-student NAS, however, can be improved.

The AKD approach \cite{Liu_2020_CVPR} applies knowledge distillation at the network level, training a reinforcement learning agent to construct an efficient student network given a teacher network and a constraint and then training that student from scratch using knowledge distillation. DNA~\citet{Li_2020_CVPR} and DONNA~\citet{moons2020distilling} take a more fine-grained approach, dividing the network into a small number of blocks, each of which contain several layers. 
During knowledge distillation, they both attempt to have student blocks mimic the output of teacher blocks, but \cite{Li_2020_CVPR} samples random paths through a mix of operators in each block, whereas \citet{moons2020distilling} trains several candidate blocks with a repeated single operation for each teacher block. They then both search for an optimal set of blocks, with DNA~ \citet{Li_2020_CVPR} using a novel ranking algorithm to predict the best set of operations within each block, and then applying a traversal search, 
while \citet{moons2020distilling} train a linear model that predicts accuracy of a set of blocks and use that to guide an evolutionary search. While both methods deliver impressive results, they differ from our approach in important ways. DNA~\citet{Li_2020_CVPR} ranks each path within a block, and then use this ranking to search over the blocks, relying on the low number of blocks to accelerate search. DONNA \citet{moons2020distilling} samples and finetunes 30 models to build a linear accuracy predictor, which incurs a significant startup cost for search. In contrast, we formulate the search problem as an integer linear optimization problems that can be solved very quickly for large networks and large pool of operations.

Table~\ref{tab:related_work} compares our work to these works in detail. We increase the granularity of network acceleration, focusing on each \emph{layer} individually instead of blocks as done in DNA and DONNA. The main advantage of focusing on layers is that it allows us to accelerate the teacher by simply \textit{replacing} inefficient layers whereas blockwise algorithms such as DNA and DONNA require \textit{searching} for an efficient subnetwork that mimics the whole block. The blockwise search introduces additional constraints. For example, both DNA and DONNA enumerate over different depth values (multiplying the search space) while we reduce depth simply using an identity operation. Additionally, DONNA assumes that the same layer in each block is repeated whereas we have more expressivity by assigning different operation to different layers.  The expressivity can be seen from the design space size in Table~\ref{tab:related_work} in which our search space is orders of magnitude larger.
On the other hand, this extremely large space necessitate the development of a highly efficient search method based on integer linear optimization (presented in Section~\ref{sec:method}). As we can see from Table~\ref{tab:related_work}, even with significantly larger search space, our total cost is lower than prior work. We additionally introduce zero-shot formation when the teacher cell and the identity operations participate in ILP, reducing the pretraining and the search costs to a minimum.

\section{Method}
\label{sec:method}

\begin{figure*}[ht!]
\vspace{-0.0cm}\centering
\begin{subfigure}[t]{.4\textwidth}
\includegraphics[width=1.\textwidth]{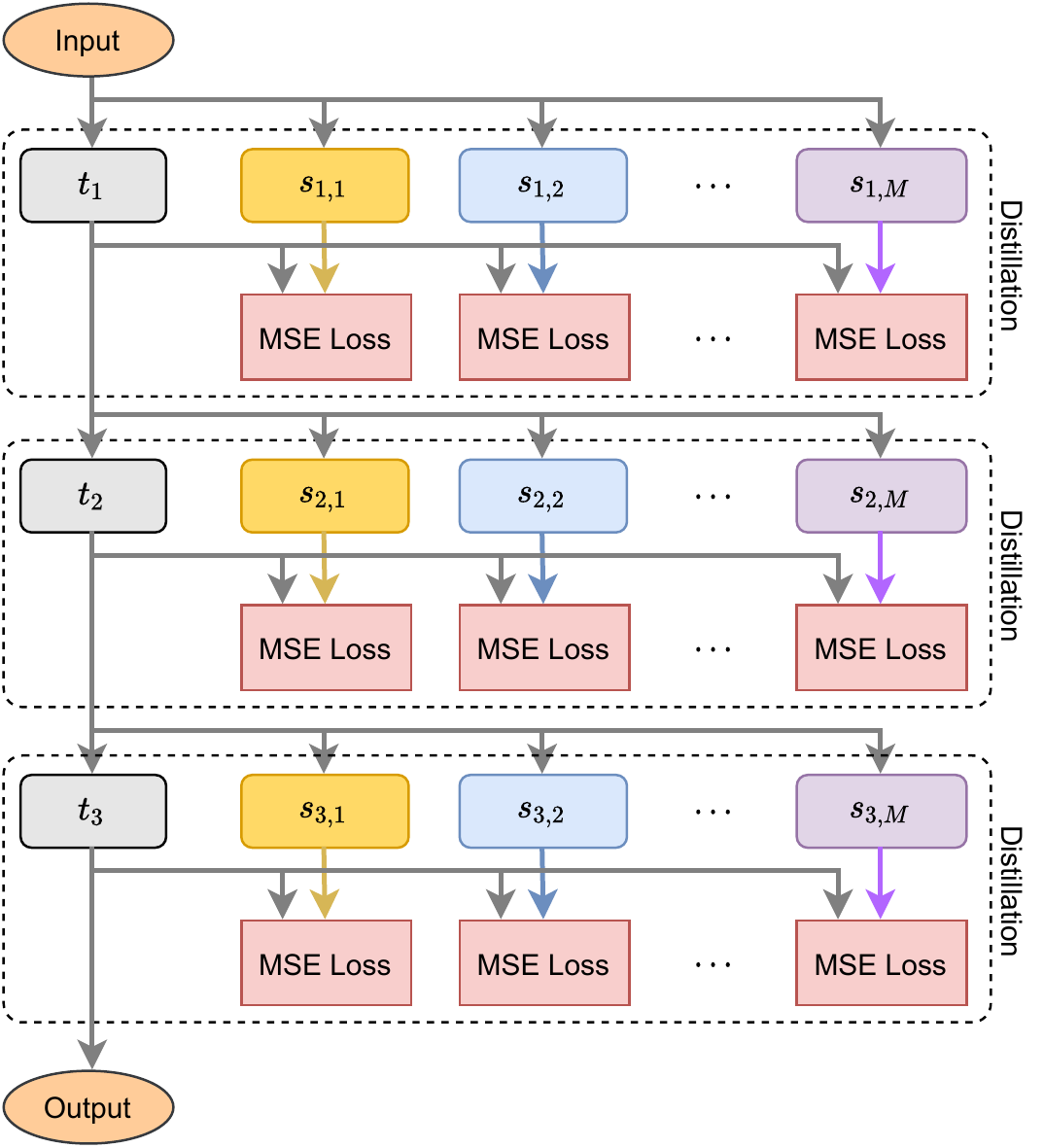}
\caption{\textbf{Candidate pretraining phase:} We minimize the MSE loss between the output of the teacher operation \(t_{i}\) and the output of each student operation on each layer \(s_{i,j}\), where the input to each operation is the teacher output from the previous layer.}
\end{subfigure}
\hspace{1mm}
\unskip\ \vrule\
\hspace{1mm}
\begin{subfigure}[t]{.4\textwidth}
\includegraphics[width=0.95\textwidth]{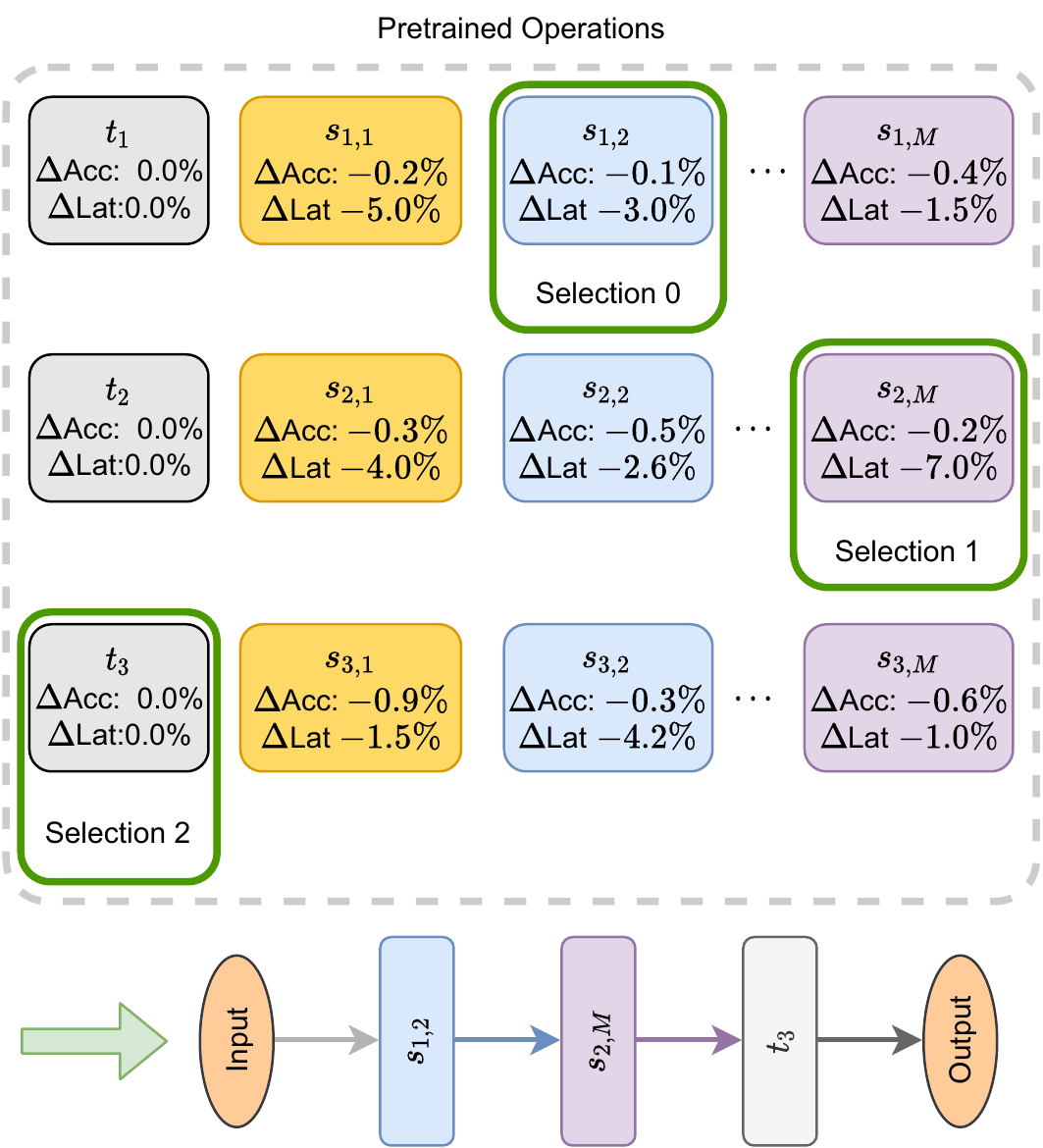}
\caption{\textbf{Operation Selection Phase:} We estimate and record in a lookup table the reduction of network accuracy and latency from replacing a teacher operation with one of the student operations. We then apply integer programming to minimize the accuracy reduction while attaining a target latency reduction.}
\end{subfigure}
\vspace{0.5cm}
\caption{\small LANA consists of two phases: a candidate pretraining phase (a) and an architecture search phase (b).\label{fig:method}}
\end{figure*}

Our goal in this paper is to accelerate a given pre-trained teacher network by replacing its inefficient operations with more efficient alternatives. Our method, visualized in Fig.~\ref{fig:method}, is composed of two phases: (i) \emph{Candidate pretraining phase} (Sec.~\ref{sec:pretraining}), in which we use distillation to train a large set of operations to approximate different layers in the original teacher architecture; (ii) \emph{Operation selection phase} (Sec.~\ref{sec:search}), in which we search for an architecture composed of a combination of the original teacher layers and pretrained efficient operations via linear optimization.

\subsection{Candidate Pretraining Phase} \label{sec:pretraining}

We represent the teacher network as the composition of $N$ teacher operations by \(\mathcal{T}\left(x\right) = t_{N} \circ t_{N - 1 } \circ \ldots \circ t_{1} \left(x\right)\), where $x$ is the input tensor, $t_{i}$ is the $i^\textrm{th}$ operation (i.e., layer) in the network.
We then define the set of \emph{candidate student operations} \(\bigcup_{i=1}^{N}\left\{s_{ij}\right\}_{j=1}^{M}\), which will be used to approximate the teacher operations. Here, $M$ denotes the number of candidate operations per layer. The student operations can draw from a wide variety of operations
-- the only requirement is that all candidate operations for a given layer must have the same input and output tensor dimensions as the teacher operation \(t_{i}\). 
We denote all the parameters (e.g., trainable convolutional filters) of the operations as $\mathbf{W} = \{w_{ij}\}_{i,j}^{N,M}$, where $w_{ij}$ denotes the parameters of the student operation $s_{ij}$. We use a set of binary vectors $\mathbf{Z} = \left\{\mathbf{z}_i\right\}_{i=1}^{N}$, where $\mathbf{z}_{i}=\{0, 1\}^M$ is a one-hot vector, to represent operation selection parameters. We denote the candidate network architecture specified by $\mathbf{Z}$ using $\mathcal{S}(x; \mathbf{Z}, \mathbf{W})$. 

The problem of optimal selection of operations is often tackled in NAS. This problem is usually formulated as a bi-level optimization that selects operations and optimizes their weights jointly \citep{liu2018darts, zoph2016neural}.
Finding the optimal architecture in hardware-aware NAS reduces to: 
\begin{align}
   \min_{\mathbf{Z}} \min_{\mathbf{W}} \quad & \underset{\textrm{objective}}{ \sum_{(x, y) \in X_{tr}}\mathcal{L}\big(\mathcal{S}(x; \mathbf{Z}, \mathbf{W}), y\big)}, &\label{eq:nas_formulation} \\
\textrm{s.t.} \quad & \underset{\textrm{budget constraint}}{\sum_{i=1}^N \mathbf{b}_{i}^T \mathbf{z}_i \leq \mathcal{B}}\ ; 
 \quad  \underset{\textrm{one op per layer}}{\mathbf{1}^T \mathbf{z}_i = 1\ \forall\ i\in[1..N]} \nonumber
\end{align}
where $\mathbf{b}_{i} \in \mathbb{R}_+^M$ is a vector of corresponding cost of each student operation (latency, number of parameters, FLOPs, etc.) in layer~\(i\). The total budget constraint is defined via scalar $\mathcal{B}$. The objective is to minimize the loss function $\mathcal{L}$ that estimates the error with respect to the correct output $y$ while meeting a budget constraint. 
In general, the optimization problem in Eq.~\ref{eq:nas_formulation} is an NP-hard combinatorial problem with an exponentially large state space (i.e., $M^N$). 
The existing NAS approaches often solve this optimization using evolutionary search~\citep{real2017large}, reinforcement learning~\citep{zoph2016neural} or differentiable search~\citep{liu2018darts}. 

However, the goal of NAS is to find an architecture in the whole search space from scratch, whereas our goal is to improve efficiency of a given teacher network by replacing operations. Thus, our search can be considered as searching in the architecture space in the proximity of the teacher network. That is why we assume that the functionality of each candidate operation is also similar to the teacher's operation, and we train each candidate operation to mimic the teacher operation 

using layer-wise feature map distillation with the mean squared error (MSE) loss:
\begin{equation} \label{eq:mse}
    \min_{\mathbf{W}} \sum_{x \in X_\textrm{tr}} \sum_{i,j}^{N,M}  \left\| t_{i}(x_{i-1}) - s_{ij}(x_{i-1}; w_{ij}) \right\|_2^2,
\end{equation}
where $X_\textrm{tr}$ is a set of training samples, and $x_{i-1}=t_{i-1}\circ t_{i-2}\circ \ldots \circ t_\text{1}\left(x\right)$ is the output of the previous layer of the teacher, fed to both the teacher and student operations. 

Our layer-wise pretraining has several advantages. First, the minimization in Eq.~\ref{eq:mse} can be decomposed into $N \times M$ independent minimization problems as $w_{i,j}$ is specific to one minimization problem per operation and layer. This allows us to train all candidate operations simultaneously in parallel. Second, since each candidate operation is tasked with an easy problem of approximating \textit{one} layer in the teacher network, we can train the student operation quickly in one epoch. In this paper, instead of solving all $N \times M$ problems in separate processes, we train a single operation for each layer in the same forward pass of the teacher to maximize reusing the output features produced in all the teacher layers. This way the pretraining phase roughly takes \(O(M)\) epochs of training a full network.

\subsection{Operation Selection Phase} \label{sec:search}
Since our goal in search is to discover an efficient network in the proximity of the teacher network, we propose a simple linear relaxation of candidate architecture loss using 
\begin{equation}
    \sum_{ X_{tr}}\mathcal{L}\big(\mathcal{S}(x; \mathbf{Z}), y\big) \approx \sum_{X_{tr}}\mathcal{L}\big(\mathcal{T}(x), y\big) + \sum_{i=1}^N \mathbf{a}_{i}^T \mathbf{z}_i,
\end{equation}
where the first term denotes the training loss of teacher which is constant and $\mathbf{a}_{i}$ is a vector of change values in the training loss per operation for layer \(i\). Our approximation bears similarity to the first-degree Taylor expansion of the student loss with the teacher as the reference point (since the teacher architecture is a member of the search space). To compute $\{\mathbf{a}_{i}\}_{i}^N$, after pretraining operations in the first stage, we plug each candidate operation one-by-one in the teacher network and we measure the change on training loss on a small labeled set. Our approximation relaxes the non-linear loss to a linear function. Although this is a weak approximation that ignores how different layers influence the final loss together, we empirically observe that it performs well in practice as a proxy for searching the student. 

Approximating the architecture loss with a linear function allows us to formulate the search problem as solving an integer linear program (ILP). This has several main advantages: (i) Although solving integer linear programs is generally NP-hard, there exist many off-the-shelf libraries that can obtain a high-quality solutions in a few seconds.  (ii) Since integer linear optimization libraries easily scale up to millions of variables, our search also scales up easily to very large number of candidate operations per layer.  (iii) We can easily formulate the search problem such that instead of one architecture, we obtain a set of diverse candidate architectures.
Formally, we denote the $k^\text{th}$ solution with \(\left\{\mathbf{Z}^{\left(k\right)}\right\}_{k=1}^{K}\), which is obtained by solving:
\begin{align}      
\min_{\textbf{Z}^{(k)}} \underset{\textrm{objective}}{\sum_{i=1}^N \mathbf{a}_{i}^T \mathbf{z}^{(k)}_i}, \
 \textrm{s.t.} & \ \underset{\textrm{budget constraint}}{\sum_{i=1}^N \mathbf{b}_{i}^T \mathbf{z}_{i}^{\left(k\right)}\leq \mathcal{B}};
\ \underset{\textrm{one op per layer}}{\mathbf{1}^T \mathbf{z}_{i}^{\left(k\right)} = 1\ \forall\ i}; \nonumber \\
& \ \underset{\textrm{overlap constraint}}{\sum_{i=1}^N {\mathbf{z}_{i}^{(k)}}^{T} \mathbf{z}^{(k')}_i} \leq \mathcal{O}, \forall k' < k  \label{eq:integer} 
\end{align}
where we minimize the change in the loss while satisfying the budget and overlap constraint.
The scalar $\mathcal{O}$ sets the maximum overlap with any previous solution which is set to \(0.7N\) in our case. We obtain $K$ diverse solutions by solving the minimization above $K$ times.

\textbf{Solving the integer linear program (ILP).} 
We use the off-the-shelf PuLP Python package to find feasible candidate solutions. 
The cost of finding the first solution is very small, often less than 1 CPU-second. As \(K\) increases, so does the difficulty of finding a feasible solution. We limit \(K\) to $\sim$100.

\textbf{Candidate architecture evaluation.} Solving Eq.~\ref{eq:integer} provides us with $K$ architectures. The linear proxy used for candidates loss is calculated in an isolated setting for each operation. To reduce the approximation error, we evaluate all $K$ architectures with pretrained weights from phase one on a small part of the training set (6k images on ImageNet) and select the architecture with the lowest loss. 

\textbf{Candidate architecture fine-tuning.} After selecting the best architecture among the $K$ candidate architectures, we fine-tune it for 100 epochs using the original objective used for training the teacher. Additionally, we add the distillation loss from teacher to student during fine-tuning. 

\vspace{-0.1cm}
\section{Experiments}
\vspace{-0.1cm}
\label{sec:experiments}

We apply LANA to the family of EfficientNetV1~\citep{tan2019efficientnet}, EfficientNetV2~\citep{tan2021efficientnetv2} and ResNeST50~\citep{zhang2020resnest}. When naming our models, we use the latency reduction ratio compared to the original model according to latency look-up table (LUT). For example, 0.25$\times$B6 indicates 4$\times$ target speedup for the B6 model.

For experiments, ImageNet-1K~\citep{ILSVRC15} is used for pretraining (1 epoch), candidate evaluation (6k training images) and finetuning (100 epochs).

We use the NVIDIA V100 GPU and Intel Xeon Silver 4114 CPU as our target hardware. A hardware specific look-up table is precomputed for each candidate operation (vectors $\mathbf{b}_i$ in Eq.~\ref{eq:integer}). 
We measure latency in 2 settings: (i) in Pytorch framework, and (ii) TensorRT ~\citep{tensorrt}. The latter performs kernel fusion for additional model optimization making it even harder to accelerate models. Note that the exact same setup is used for evaluating latency of \textbf{all} competing models, our models, and baselines.  Actual latency on the target platform is reported in the tables.  

\textbf{Candidate operations. } 
We construct a large of pool of diverse candidate operation including $M=197$ operations for each layer of teacher. 
Our operations include:

{\parskip=2pt
\begin{itemize}[topsep=1pt,itemsep=1pt,partopsep=0pt,parsep=0pt,leftmargin=0pt]

\item[]\underline{Teacher operation} is used as is in the pretrained model with teacher model accuracy.

\item[]\underline{Identity} is used to skip teacher's operation. It changes the depth of the network. 

\item[]\underline{Inverted residual blocks} \texttt{efn} \citep{sandler2018mobilenetv2} and \texttt{efnv2} \citep{tan2021efficientnetv2}  with varying expansion factor $e{=}\{1,3,6\}$, squeeze and excitation ratio $se{=}\{\textrm{No}, 0.04, 0.025\}$, and kernel size $k{=}\{1, 3, 5\}$. 

\item[]\underline{Dense convolution blocks} inspired by~\cite{he2016identity} with (i) two stacked convolution (\texttt{cb\_stack}) with CBRCB structure, C-conv, B-batchnorm, R-Relu; (ii) bottleneck architecture (\texttt{cb\_bottle}) with CBR-CBR-CB; (ii) CB pair (\texttt{cb\_res}); (iii) RepVGG block~\citep{ding2021repvgg}; (iv) CBR pairs with perturbations as \texttt{conv\_cs}. For all models we vary kernel size $k = \{1,3,5,7\}$ and width $w = \{1/16, 1/10, 1/8, 1/5, 1/4, 1/2, 1, 2, 3, 4\}$.

\item[]\underline{Transformer variations} (i) visual transformer block~(\texttt{vit}) \citep{dosovitskiy2020image} with depth $d=\{1,2\}$, dimension $w=\{2^5, 2^6, 2^7, 2^8, 2^9, 2^{10}\}$ and heads $h=\{4,8,16\}$; (ii) bottleneck transformers~\citep{srinivas2021bottleneck} with 4 heads and expansion factor $e = \{1/4, 1/2, 1, 2, 3, 4\}$; (iii) lambda bottleneck layers~\citep{bello2021lambdanetworks} with expansion $e=\{1/4, 1/2, 1, 2, 3, 4\}$. 
\end{itemize}
}

\begin{table}[t]
\centering
    \renewcommand{\arraystretch}{0.65}
    \setlength{\tabcolsep}{3pt}
    \centering
        \caption{Models optimized with LANA for GPU inference, evaluated on ImageNet-1K. Latency is computed for a batch of 128 images over 10 runs on the actual hardware. Models are grouped by the latency to demonstrate accuracy improvement over the vanilla EfficientNet models with default model scaling.}
        
    \resizebox{0.49\textwidth}{!}{%
    \begin{tabular}{p{2.4cm}u{1.4cm}x{1.1cm}lcc}
    \toprule
    \multirow{2}{*}{Method} & \multirow{2}{*}{Variant} & Res & Accuracy & \multicolumn{2}{c}{Latency(ms)} \\
     & & {px} & (\%)  & TensorRT & PyTorch\\
    \midrule
    \midrule
    \multicolumn{6}{c}{\textbf{EfficientNetV1}}  \\
    \midrule
    \midrule
    EfficientNetV1 & 1.00xB0         & 224  & 77.70 & 17.9 & 35.6 \\
    LANA  & 0.45xB2                & 260  & \textbf{79.71 (+2.01)} & \textbf{16.2} & \textbf{30.2}\\      
    \midrule
    EfficientNetV1 & 1.00xB1         & 240  & 78.83 & 29.3 & 59.0 \\
    LANA & 0.55xB2                  & 260  & 80.11 (+1.28) & \textbf{20.6} & \textbf{48.7}\\      
    LANA & 0.20xB4                & 380  & 80.33 (+1.50) & 33.0 & 52.2\\
    LANA & 0.25xB4                 & 380  & \textbf{81.83 (+3.00)} & 30.4 & 64.5\\
    \midrule
    EfficientNetV1 & 1.0xB2         & 260  & 80.07 & 38.2 & 77.1 \\
    LANA & 0.3xB4                 & 380  & \textbf{82.16 (+2.09)} & \textbf{38.8} & 81.8\\        
    \midrule
    EfficientNetV1 & 1.0xB3         & 300  & 81.67 & 67.2 & 125.9 \\
    LANA  & 0.5xB4                & 380  & \textbf{82.66 (+0.99)} & \textbf{61.4} & 148.1\\            
    \midrule
    EfficientNetV1 & 1.00xB4         & 380  & 83.02 & 132.0 & 262.4  \\
    LANA & 0.25xB6               & 528  & \textbf{83.77 (+0.75)} &  \textbf{128.8} & 282.1\\
    \midrule
    EfficientNetV1 & 1.0xB5         & 456  & 83.81 & 265.7 & 525.6  \\
    LANA &0.5xB6               & 528  & \textbf{83.99 (+0.18)} & \textbf{266.5} & 561.2 \\
    \midrule
    EfficientNetV1 & 1.0xB6         & 528  & 84.11 & 466.7 & 895.2  \\
        \midrule
    \midrule
    \multicolumn{6}{c}{\textbf{EfficientNetV2}}  \\
    \midrule
    \midrule
    EfficientNetV2 & 1.00xB1         & 240  & 79.46 & 17.9 & 44.7 \\
    LANA & 0.45xB3                  & 300  & \textbf{80.30 (+0.84)} & \textbf{17.8} & \textbf{43.0}\\      
    \midrule
    EfficientNetV2 & 1.0xB2         & 260  & 80.21 & 24.3 & 58.9 \\
    LANA & 0.6xB3                    & 300  & \textbf{81.14 (+0.93)} & \textbf{23.8} & \textbf{56.1}\\      
    \midrule
    EfficientNetV2 & 1.0xB3       & 300  & 81.97 & 41.2 & 91.6 \\
    \midrule
    \midrule
    \multicolumn{6}{c}{\textbf{ResNeST50d\_1s4x24d}}  \\
    \midrule
    \midrule
    ResNeST50 & 1.0x       & 224  & 80.99 & 32.3 & 74.0 \\
    LANA & 0.7x      & 224  & 80.85 & \textbf{22.3(1.45x)} & \textbf{52.7} \\
    \bottomrule
    \end{tabular}
    }%
    
     \centering
     \includegraphics[width=0.45\textwidth, trim={0cm 0 0cm 0cm},clip]{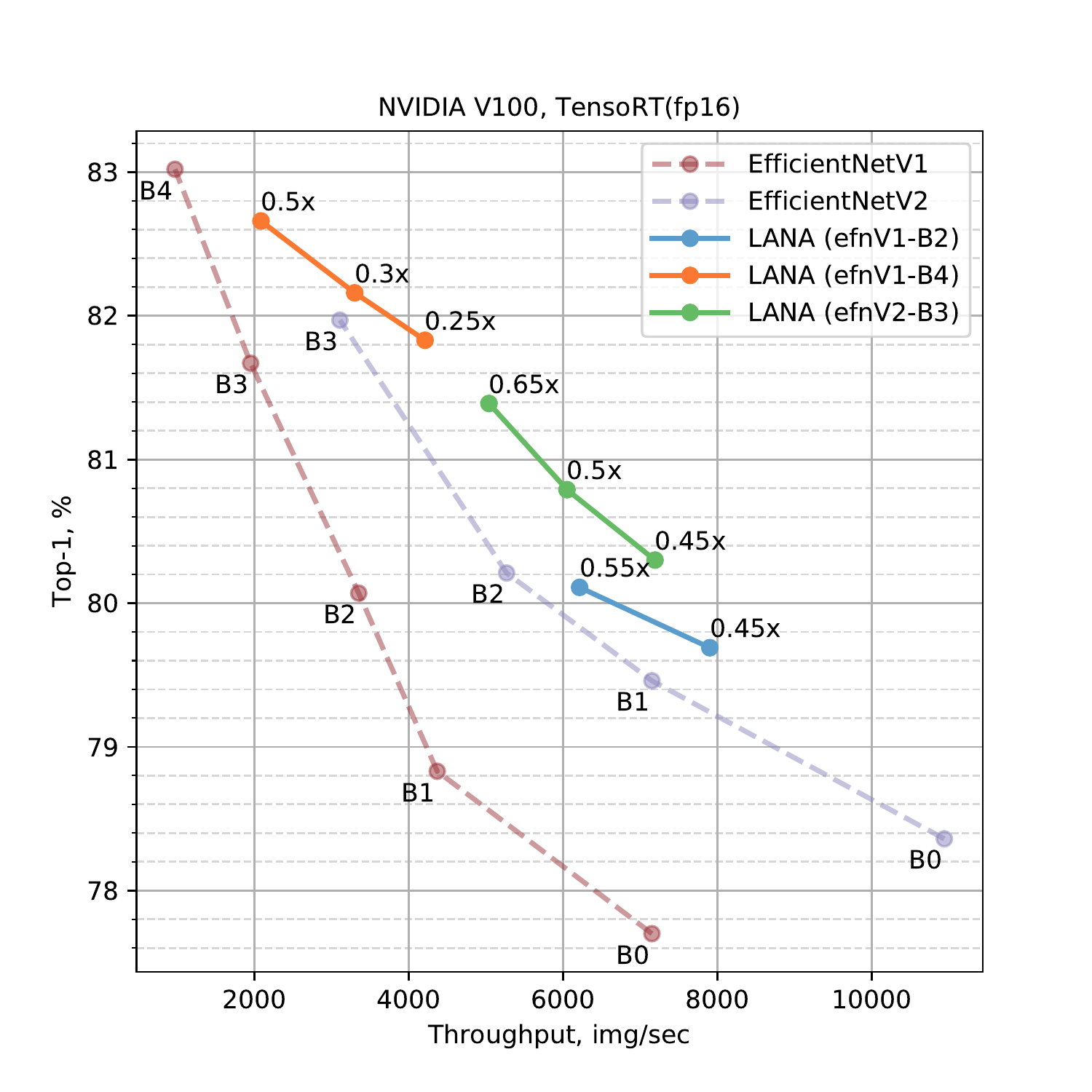}
     \captionof{figure}{A visual summary of the EfficientNetV1 and EfficientNetV2 model families accelerated by LANA. Accelerated EfficientNetV1 models outperform EfficientNetV2 models.}
     \label{fig:imagenet_comparison}
    \label{tab:imagenet_comparison}
\end{table}

With the pool of 197 operations, distilling from an EfficientNet-B6 model with $46$ layers yields a design space of the size \(197^{46}\!\approx\!10^{100}\).

\subsection{EfficientNet and ResNeST Derivatives}

Our experimental results on accelerating EfficientNetV1(B2, B4, B6), EfficientNetV2(B3), and ResNeST50 family for GPUs are shown in Table~\ref{tab:imagenet_comparison} and Figure~\ref{fig:imagenet_comparison}.
Comparison with more models from \texttt{timm} is in the Appendix. 

\textbf{Results demonstrate that: }
{\parskip=2pt
\begin{itemize}[topsep=1pt,itemsep=1pt,partopsep=0pt,parsep=0pt,leftmargin=10pt]
\item LANA achieves an \textbf{accuracy boost of} $\mathbf{0.18-3.0}\%$ for all models  when compressing larger models to the latency level of smaller models (see EfficientNet models and the corresponding LANA models in the same latency group).
\item LANA achieves \textbf{significant real speed-ups} with little to no accuracy drop: 
(i) Efficiei87ntNetV1-B6 accelerated by 3.6x and 1.8x times by trading-off 0.34\% and 0.11\% accuracy, respectively; (ii) B2 variant is accelerated 2.4x and 1.9x times with 0.36\% and no accuracy loss, respectively. (iii) ResNeST50 is accelerated 1.5x with 0.84\% accuracy drop.
\end{itemize}
}

A detailed look on EfficientNets is shown in Figure~\ref{fig:imagenet_comparison}, where we observe that EfficientNetV1 models are accelerated beyond EfficientNetV2. LANA generates models that have better accuracy-throughput trade-off when comparing models under the same accuracy or the same latency. LANA also allows us to optimize models for different hardware at a little cost. Only a new LUT is required to get optimal model for a new hardware without pretraining the candidate operations again. We present models optimized for CPU in Table~\ref{tab:efficientnetv1_cpu} which are obtained using a different LUT.  




 \begin{table}[b]
  \centering
  \caption{EfficientNetV1 accelerated by LANA for CPU inference.}
  \vspace{-0.2cm}
    \resizebox{0.4\textwidth}{!}{%
    \begin{tabular}{lclc}
        \toprule
        \multirow{2}{*}{Model} & Res. & Accuracy & Latency \\
         & (px) & (\%) & Pytorch (ms) \\
        \midrule
        \midrule
        EfficientNetB0          & 224  & 77.70 & 57\\
        0.4xB2 (Xeon)           & 260 & 78.11 (+0.97) & 48 \\
        0.5xB2 (Xeon)           & 260 & 78.87 (+1.17) & 58  \\
        \midrule
        EfficientNetB1          & 240 & 78.83 & 86\\
        0.7xB2 (Xeon)           & 260 & 79.89 (+1.06) & 80 \\
        \midrule
        EfficientNetB2          & 260 & 80.07 & 113\\
        \bottomrule
    \end{tabular}
    }%
  \label{tab:efficientnetv1_cpu}
 \end{table}

\subsection{Analysis}


Here, we provide detailed ablations to analyze our design choices in LANA for both pretraining and search phases, along with observed insights. Unless otherwise stated, we used EfficientNetV2-B3 as our teacher for the ablation.

\begin{figure*}
     \centering
         \begin{subfigure}[b]{0.30\textwidth}
         \centering
         \includegraphics[width=1.\textwidth,trim={0 0 0px 0},clip]{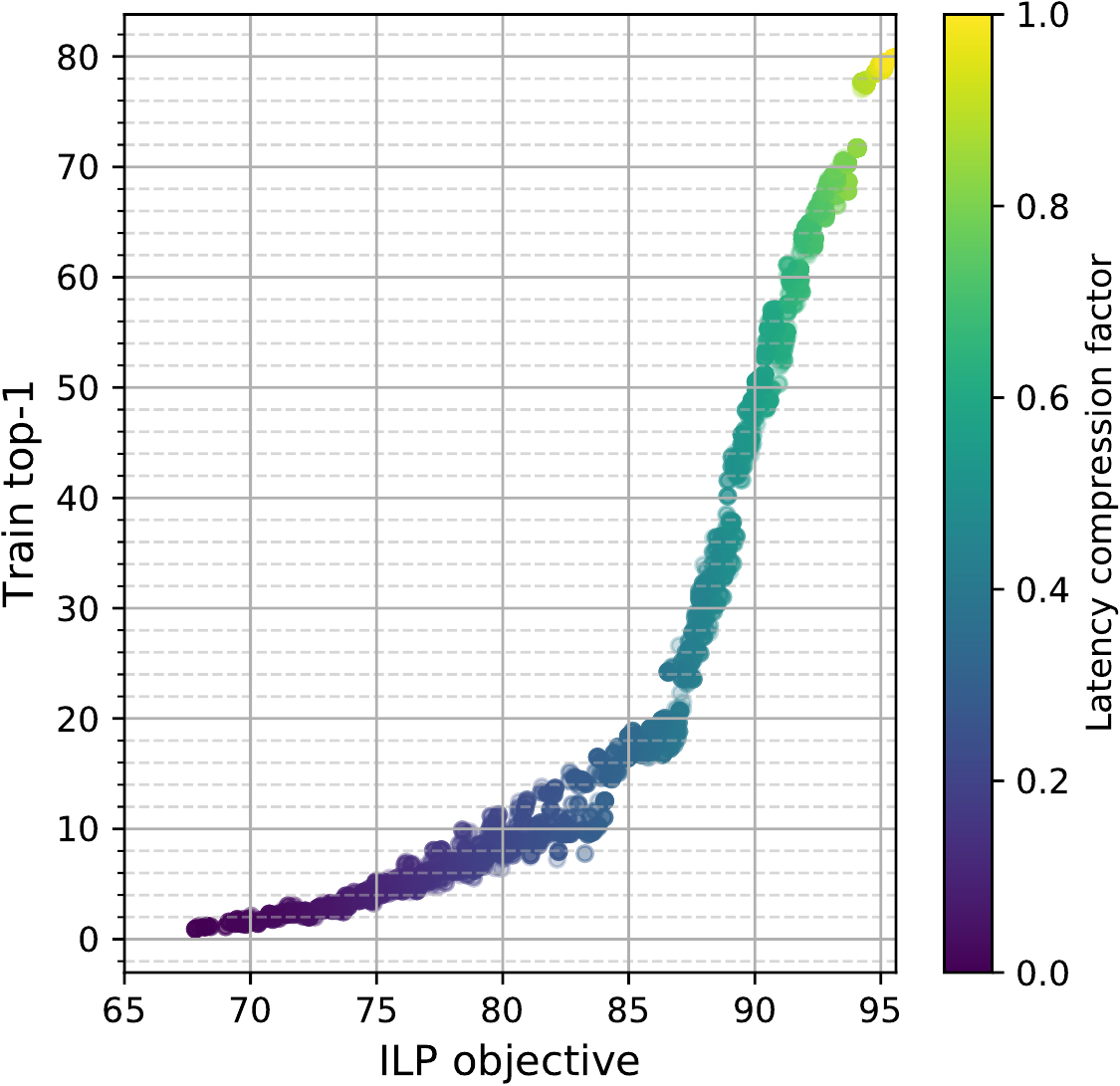}
         \caption{Optimization objective}
         \label{fig:linear_relaxation}
     \end{subfigure}\hspace{5mm}
     \begin{subfigure}[b]{0.30\textwidth}
         \centering
         \includegraphics[width=1.\textwidth,trim={0 0cm 0px 0},clip]{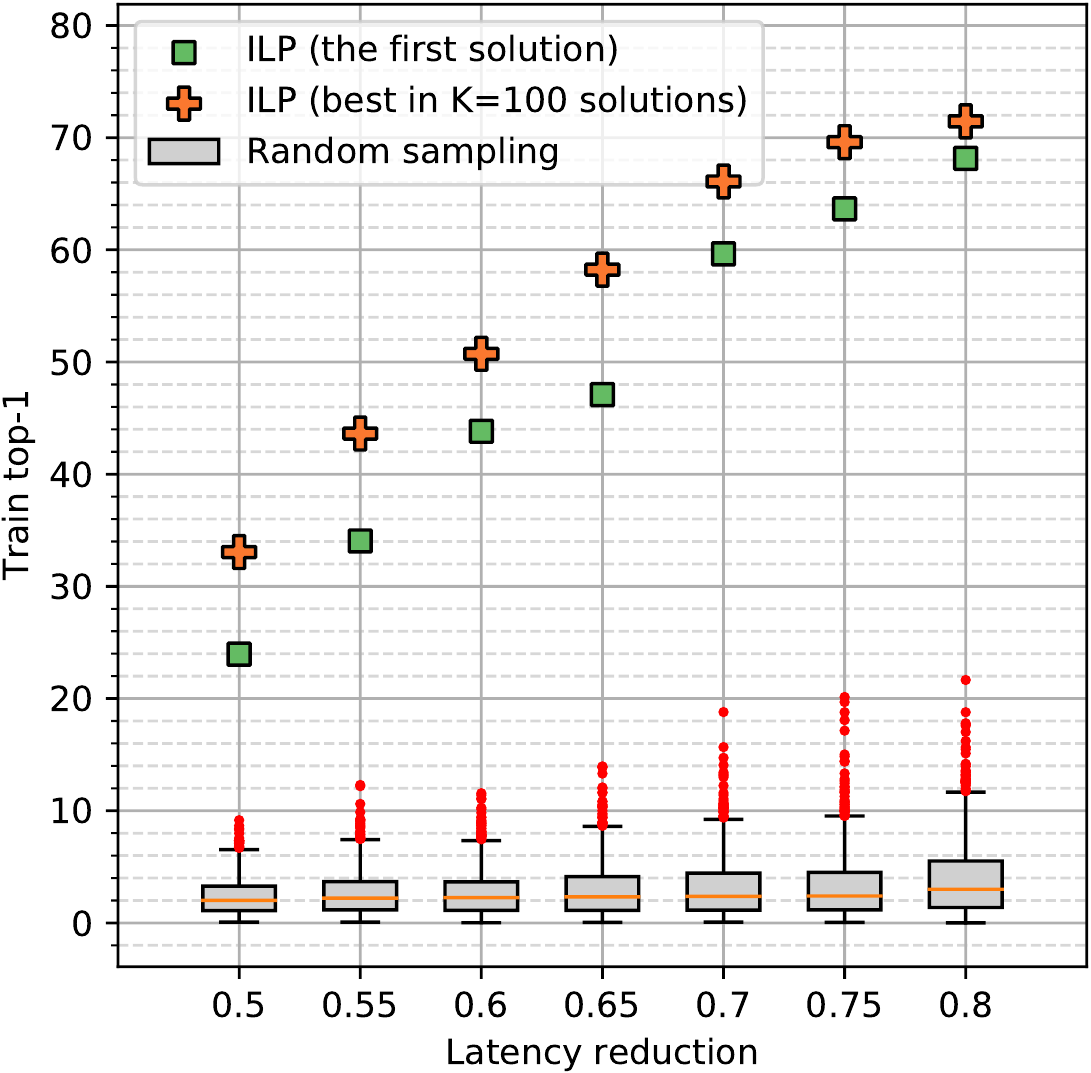}
         \caption{ILP vs random}
         \label{fig:ilp_vs_random}
     \end{subfigure}\hspace{5mm}
     \begin{subfigure}[b]{0.3\textwidth}
         \centering
         \includegraphics[width=1.\textwidth,trim={0 0 0px 0},clip]{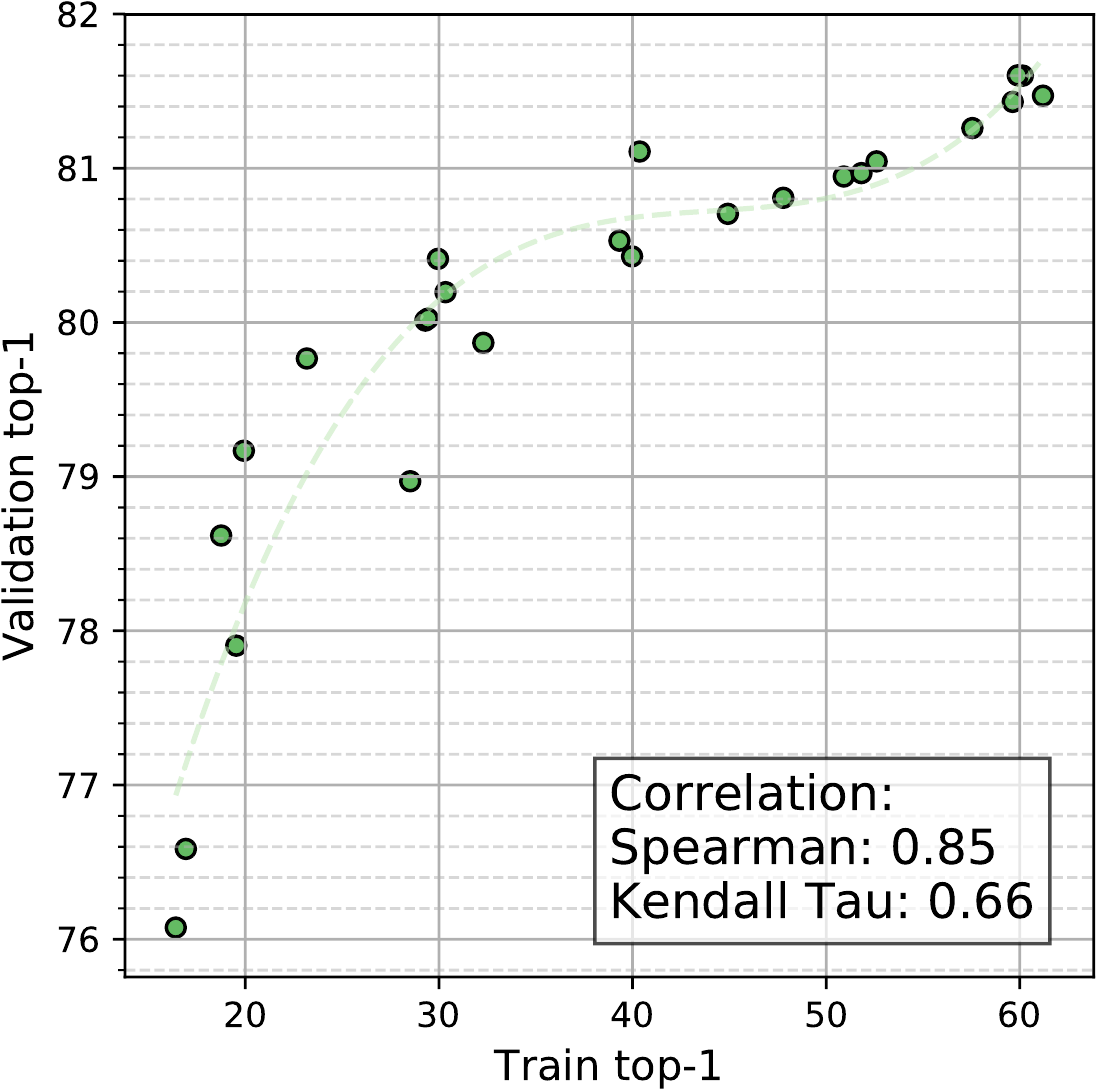}
         \caption{Candidate evaluation}
         \label{fig:candidate_evaluation}
     \end{subfigure}
     \vspace{6mm}
        \caption{Analyzing ILP performance on EfficientNetV2-B3. ILP results in significantly higher model accuracy before finetuning than 1k randomly sampled architectures in (\subref{fig:ilp_vs_random}). Accuracy monotonically increases with ILP objective (\subref{fig:linear_relaxation}). Model accuracy before finetuning correctly ranks models after finetuning (\subref{fig:candidate_evaluation}). Train top-1 is measured \textit{before} finetuning, while Validation top-1 is \textit{after}.
        }
        
        \label{fig:ILP_analysis}

\end{figure*}

\textbf{Linear relaxation} in architecture search assumes that a candidate architecture can be scored by a fitness metric measured independently for all operations. 
Although this relaxation is not accurate, we observe a strong correlation between our linear objective and the training loss of the full architecture.
This assumption is verified by sampling 1000 architectures (different budget constraints), optimizing the ILP objective, and measuring the real loss function. Results are shown in Fig.~\ref{fig:ILP_analysis}\subref{fig:linear_relaxation} using the train accuracy as the loss. We observe that ILP objective ranks models with respect to the measured accuracy correctly under different latency budgets. The Kendall Tau correlation is $0.966$.

To evaluate the quality of the solution provided with ILP, we compare it with random sampling. The comparison is shown in Fig.~\ref{fig:ILP_analysis}\subref{fig:ilp_vs_random}, where we sample 1000 random architectures for 7 latency budgets. The box plots indicate the poor performance of the randomly sampled architectures. 
The first ILP solution has significantly higher accuracy than random architecture. Furthermore, finding multiple diverse solutions is possible with ILP using the overlap constraint. If we increase the number of solutions found by ILP from $K{=}1$ to $K{=}100$, performance improves further.

\textbf{Candidate architecture evaluation} plays an important role in LANA. This step finds the best architecture quickly out of the diverse candidates generated by the ILP solver, by evaluating them on 6k images from the train data. The procedure is built on the assumption that the accuracy of the model on the training data before finetuning (just by plugging all candidate operations) is a reasonable indicator of the relative model performance after finetuning. As shown in Fig.~\ref{fig:ILP_analysis}\subref{fig:ilp_vs_random}, when plugging pretrained operations into the teacher, the accuracy is high (it is above 30\%, even at an acceleration factor of 2$\times$). For EfficientNetV1, this is above 50\% for the same compression factor. 

\begin{table}[t]

\centering
\caption{\small Comparing methods for candidate selection (NAS). Our proposed ILP is better ($+0.43\%$) and 821$\times$ faster.}
\label{tab:nas_comparison}
\vspace{-0.2cm}
\resizebox{0.35\textwidth}{!}{
\begin{tabular}{lcc}
\toprule
Method                            & Accuracy & Search cost\\
\midrule
\textbf{ILP, K=100 (ours)}                                            & 79.28   & 4.5 CPU/m \\
Random, found 80 arch                                  & 76.44   & 1.4 CPU/m     \\ 
SNAS~\citep{xie2018snas}                   &  74.20 & 16.3 GPU/h  \\ 
E-NAS~\citep{pham2018efficient}   &  78.85 & 61.6 GPU/h  \\ 
\bottomrule \vspace{-0.5cm}  
\end{tabular}
}
\end{table}

\textbf{Comparing with other NAS approaches.} We compare our search algorithm with other popular approaches to solve Eq.~(\ref{eq:nas_formulation}), including: (i) \textit{Random} architecture sampling within a latency constrain; (ii) \textit{Differentiable search with Gumbel Softmax} -- a popular approach in NAS to relax binary optimization as a continuous variable optimization via learning the sampling distribution~\citep{xie2018snas, Wu_2019_CVPR,vahdat2020unas}. We follow SNAS~\citep{xie2018snas} in this experiment; (iii) \textit{REINFORCE} is a stochastic optimization framework borrowed from reinforcement learning and adopted for architecture search~\citep{pham2018efficient, zoph2018learning,tan2019mnasnet}. We follow an E-NAS-like~\citep{pham2018efficient} architecture search for (iii) and use weight sharing for (ii) and (iii).
Experiments are conducted on EfficientNetV1-B2 accelerated to 0.45$\times$ original latency. The final validation top-1 accuracy after finetuning are presented in Table~\ref{tab:nas_comparison}. Our proposed ILP achieves higher accuracy ($+0.43\%$) compared to the second best method E-NAS while being 821$\times$ faster in search. 

\begin{table}
\centering
\caption{\small Zero-shot LANA with only skip connections. 
\vspace{-0.2cm}} 
\label{tab:zero_shot_LANA}
\resizebox{0.45\textwidth}{!}{%
\begin{tabular}{lccc}
\toprule
Setup                            & \multicolumn{2}{c}{Top-1 Acc.} & Latency(ms) \\
                                 & Zero-shot &  All operations & TensorRT \\
\midrule
0.45xEfficientNetV1-B2            & 78.68 & 79.71   & 16.2\\
0.55xEfficientNetV1-B2            & 79.40 & 80.11   & 20.6\\
\midrule
EfficientNetV1-B0         & \multicolumn{2}{c}{77.70} & 17.9 \\
EfficientNetV1-B2            & \multicolumn{2}{c}{80.07} & 38.8\\
\bottomrule 
\end{tabular}
}
\end{table}

\textbf{Zero-shot LANA.} Our method can be applied without pretraining procedure if only teacher cells and \textit{Identity} (skip) operation are used ($M=2$ operations per layer). Only the vector for the change in loss for the \textit{Identity} operator will be required alongside the LUT for the teacher operations. This allows us to do zero-shot network acceleration without any pretraining as reported in Table~\ref{tab:zero_shot_LANA}. We observe that LANA efficiently finds residual blocks that can be skipped. This unique property of LANA is enabled because of accelerating layers as apposed to blocks as done in \cite{Li_2020_CVPR, moons2020distilling}.

\textbf{Pretraining insights.} 
\label{sec:pretraining_insights}
To gain more insights into the tradeoff between the accuracy and speed of each operation, we analyze the pretrained candidate operation pool for EffientNetV1B2. 
A detailed figure is shown in the appendix. Here, we provide general observations.

We observe that no operation outperforms the teacher in terms of accuracy; changing pretraining loss from per-layer MSE to full-student cross-entropy may change this but that comes with an increased costs of pretraining.
We also see that it is increasingly difficult to recover the teacher's accuracy as the depth in the network increases. The speedups achievable are roughly comparable across different depths, however, we note that achieving such speedups earlier in the network is particularly effective towards reducing total latency due to the first third of the layers accounting for 54\% of the total inference time.

Looking at individual operations, we observe that inverted residual blocks ({\small\texttt{efn}}, {\small \texttt{efnv2}}) are the most accurate throughout the network, at the expense of increased computational cost (\emph{i.e.,} lower speedups). Dense convolutions ({\small \texttt{cb\_stack}}, {\small \texttt{cb\_bottle}}, {\small \texttt{conv\_cs}}, {\small \texttt{cb\_res}}) exhibit a good compromise between accuracy and speed, with stacked convolutions being particularly effective earlier in the network.
Visual transformer blocks ({\small \texttt{ViT}}) and bottleneck transformer blocks ({\small \texttt{bot\_trans}}) show neither a speedup advantage nor are able to recover the accuracy of the teacher.


\subsubsection{Architecture insights}

In the appendices, we visualize the final architectures discovered by LANA. Next, we share the insights observed on these architectures. 


\textbf{EfficientNetV1.}
 Observing final architectures obtained by LANA on the EfficientNetV1 family, particularly the 0.55$\times$B2 version optimized for GPUs,  we discover that most of the modifications are done to the first half of the model: (i) squeeze-and-excitation is removed in the early layers; (ii) dense convolutions (like inverted stacked or bottleneck residual blocks) replace depth-wise separable counterparts; (iii) the expansion factor is reduced from 6 to 3.5 on average. \textit{Surprisingly, LANA automatically discovers the same design choices that are introduced in the EfficientNetV2 family when optimized for datacenter inference.} 

\textbf{EfficientNetV2.} LANA accelerates EfficientNetV2-B3 by 2$\times$, leading to the following conclusions: (i) the second conv-bn-act layer 
is not needed and can be removed; (ii) the second third 
of the model benefits from reducing the expansion factor from 4 to 1 without squeeze-and-excitation operations. With these simplifications, the accelerated model still outperforms EfficientNetV2-B2 and B1.

\textbf{ResNeST50.} LANA on ResNeST50d 
discovers that cardinality can be reduced from 4 to 1 or 2 for most blocks without any loss in accuracy, yielding a 1.45$\times$ speedup.

\subsubsection{Ablations on finetuning}

\hspace{1.9mm} \textbf{Pretrained weights.} We look deeper into the finetuning step. 
For this experiment, we select our 0.45$\times$EfficientNetV1-B2 with the final accuracy of 79.71\%. 
Reinitializing all weights in the model, as opposed to loading them from the pretraining stage, results in 79.42\%. The result indicates the importance of pretraining stage (i) to find a strong architecture and (ii) to boost finetuning.

\textbf{Knowledge distillation} plays a key role in student-teacher setups. When it is disabled, we observe an accuracy degradation of 0.65\%. This emphasizes the benefit of training a larger model and then distilling to a smaller one. We further verify whether we can achieve a similar high accuracy using knowledge distillation from EfficientNetV1-B2 to vanilla EfficientNetV1-B0 in the same setting. The top-1 accuracy of 78.72\% is still 1\% less than LANA's accuracy. 
When both models are trained from scratch with the distillation loss, LANA 0.45xB2 achieves 79.42\% while EfficientNetV1-B0 achieves 78.01\%. 
This verifies that results similar to LANA cannot be obtained simply by distilling knowledge from larger EfficientNets to smaller ones.

\begin{table}
    \centering
    \setlength{\tabcolsep}{3pt}

\caption{{Ablations on the length of finetuning step. On the right side, we report EfficientNet accuracy for models with the same latency as ours.} \vspace{-0.0cm}}
\vspace{-.2cm}
\label{tab:finetuning_length}
\resizebox{0.49\textwidth}{!}{%
\begin{tabular}{x{1.5cm}x{1.2cm}x{1.2cm}x{1.2cm}x{1.2cm}x{1.2cm}|x{2.0cm}}
\toprule
LANA  &  \multicolumn{5}{c|}{Epochs} & \small{EfficientNetV1} \\ \cline{2-6}
model &   5 & 10 & 25 & 50 & 100 & alternative\\ \hline
 0.45xB2  &78.69 & 79.08 & 79.19 &	79.58 &	79.71 & 77.70 (B0) \\ 
 0.55xB2  & 79.05 & 79.47 & 79.84 &	80.00 &	80.11 & 78.83 (B1)\\
\bottomrule
\end{tabular}}
\end{table}

\textbf{Length of finetuning.} Pretrained operations have already been trained to mimic the teacher layer. Therefore, even before finetuning the student model can be already adept at the task. Next, we evaluate how does the length of finetuning affects the final accuracy in Table~\ref{tab:finetuning_length}. Even with only 5 epochs LANA outperforms the vanilla EfficientNet counterparts.

\begin{table}
    \centering
    \setlength{\tabcolsep}{3pt}

\caption{Impact of the search space on 0.55$\times$EfficientNetV1-B2 compression. Two operations correspond to Zero-shot LANA.}
\label{tab:search_space}
    \vspace{-0.2cm}
\resizebox{0.39\textwidth}{!}{%
\begin{tabular}{lx{0.07\textwidth}x{0.08\textwidth}x{0.08\textwidth}x{0.08\textwidth}}
\toprule
Operations \# & 2 & 5 & 10 & All \\
\midrule
Space size   & $O(10^{7})$ &  $O(10^{16})$ & $O(10^{23})$ & $O(10^{46})$ \\
Accuracy  & 79.40 &	79.52 &	79.66 &	80.00 \\

\bottomrule

\end{tabular}}
\end{table}

\textbf{Search space size.} ILP enables us to perform NAS in a very large space ($O(10^{100})$). To verify the benefit of large search space, we experiment with a restricted search space. For this, we randomly pick 2, 5 and 10 operations per layer to participate in search and finetuning for 50 epochs. We observe clear improvements from increasing the search space, shown Table~\ref{tab:search_space} (results are averaged over 5 runs).   

\begin{table}[t]
  \caption{Comparison to \cite{moons2020distilling}. Latency values measured on V100 GPU at batch size $32$ with PyTorch FP32. All models were finetuned for 50 epochs only. }
  \label{tab:donna_comparison}
  \vspace{-0.2cm}
  \centering
  \resizebox{0.4\textwidth}{!}{
  \begin{tabular}{lcc}
    \toprule
    \multicolumn{1}{c}{Models}   & Latency    & Top-1 Acc.     \\
             & (ms) $\downarrow$ &  ($\%$) $\uparrow$ \\
    \midrule
    DONNA~\cite{moons2020distilling} & $20.0$ & $78.9$ \\
    \textbf{LANA} (0.45xEfficientNetV1-B2)  & \textbf{18.9} & \textbf{79.58 (+0.68)}  \\
    \midrule
    DONNA~\cite{moons2020distilling} & 25.0 & 79.5 \\
    \textbf{LANA} (0.55xEfficientNetV1-B2) & \textbf{24.1} & \textbf{80.0 (+0.5)}\\
    \bottomrule
  \end{tabular}
  }
  \vspace{-0pt}
\end{table}

\textbf{Comparison to~\cite{moons2020distilling}.} DONNA has a similar motivation ours and has reported results comparable to our setup. We observe significant improvements over DONNA in the Table~\ref{tab:donna_comparison} with lower latency and higher accuracy.

\section{Conclusion}
In this paper, we proposed LANA, a hardware-aware network transformation framework for accelerating pretrained neural networks. LANA uses a NAS-like search to replace inefficient operations with more efficient alternatives. It tackles this problem in two phases including a candidate pretraining phase and a search phase. The availability of the teacher network allows us to estimate the change in accuracy for each operation at each layer. Using this, we formulate the search problem as solving a linear integer optimization problem, which outperforms the commonly used NAS algorithms while being orders of magnitude faster. We applied our framework to accelerate EfficientNets (V1 and V2) and ResNets with a pool of 197 operations per layer and we observed that LANA accelerates these architectures by several folds with only a small drop in the accuracy. We analyzed the selected operations and we provided new insights for future neural architecture designs.

\textbf{Limitations.}
The student performance in LANA is bounded by the teacher, and it rarely passes the teacher in terms of accuracy. Additionally, the output dimension of layers in the student can not be changed, and must remain the same as in the original teacher. 

\textbf{Future work.} We envision that a layer-wise network acceleration framework like LANA can host a wide range of automatically and manually designed network operations, developed in the community. Users with little knowledge of network architectures can then import their inefficient networks in the framework and use LANA to accelerate them for their target hardware. Our design principals in LANA consisting of extremely large operation pool, efficient layer-wise pretraining, and lightening fast search help us realize this vision. We will release the source code for LANA publicly to facilitate future development in this space.

\section*{Acknowledgement}
This work was conducted jointly at NVIDIA and Microsoft research. The authors would like to thank Azure ML and the office of the CTO at Microsoft as well as the NVIDIA NGC team for compute support.

{
    \small
    \bibliographystyle{ieee_fullname}
    \bibliography{references}
}

\newpage
\section*{Appendix}
\label{sec:appendix}

\begin{figure*}[h]
     \centering
     \begin{subfigure}[b]{0.32\textwidth}
         \centering
         \includegraphics[width=\textwidth]{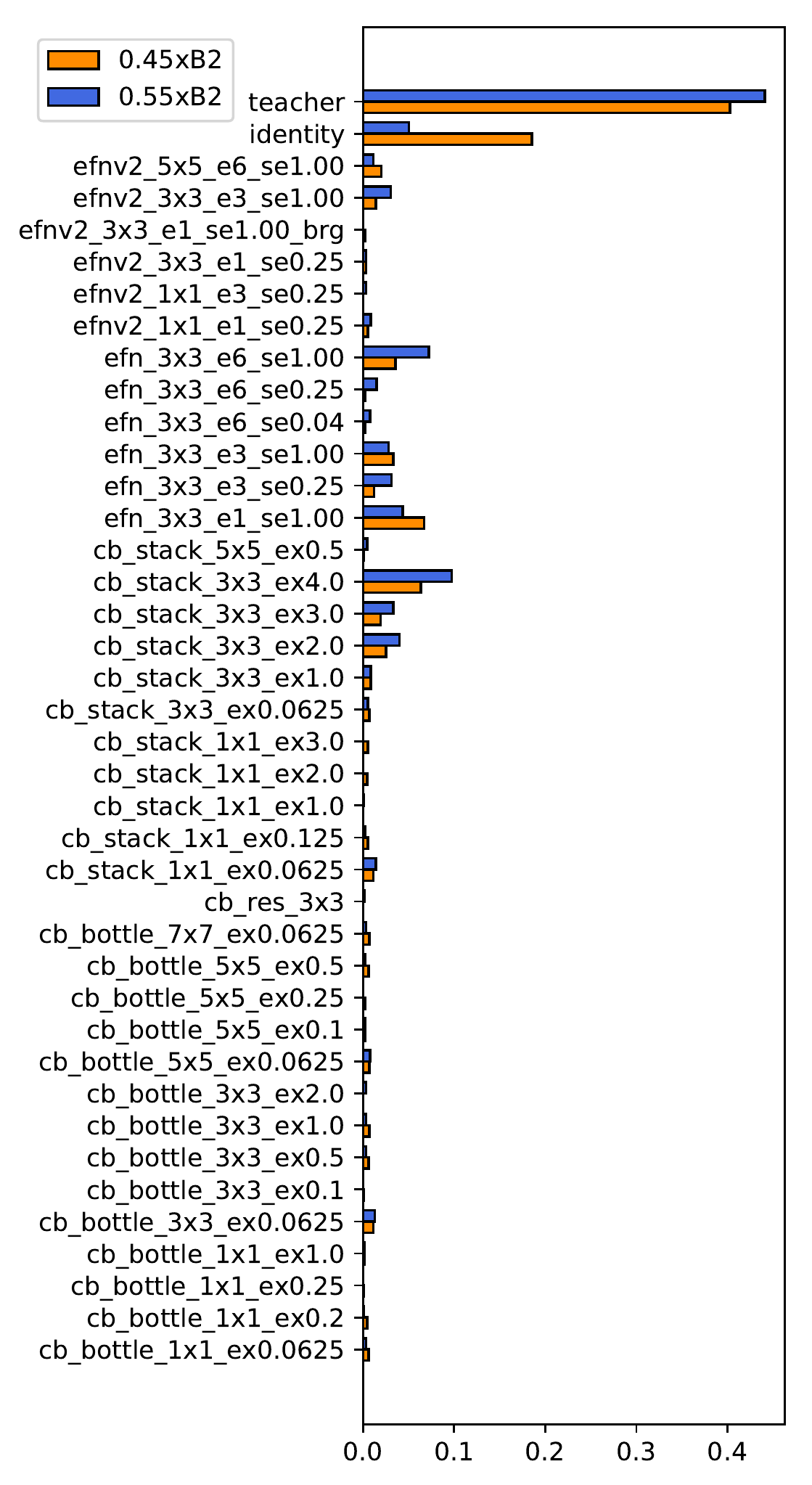}
         \caption{B2 selected ops}
         \label{fig:ops_b2}
     \end{subfigure}
     \hfill
     \begin{subfigure}[b]{0.32\textwidth}
         \centering
         \includegraphics[width=\textwidth]{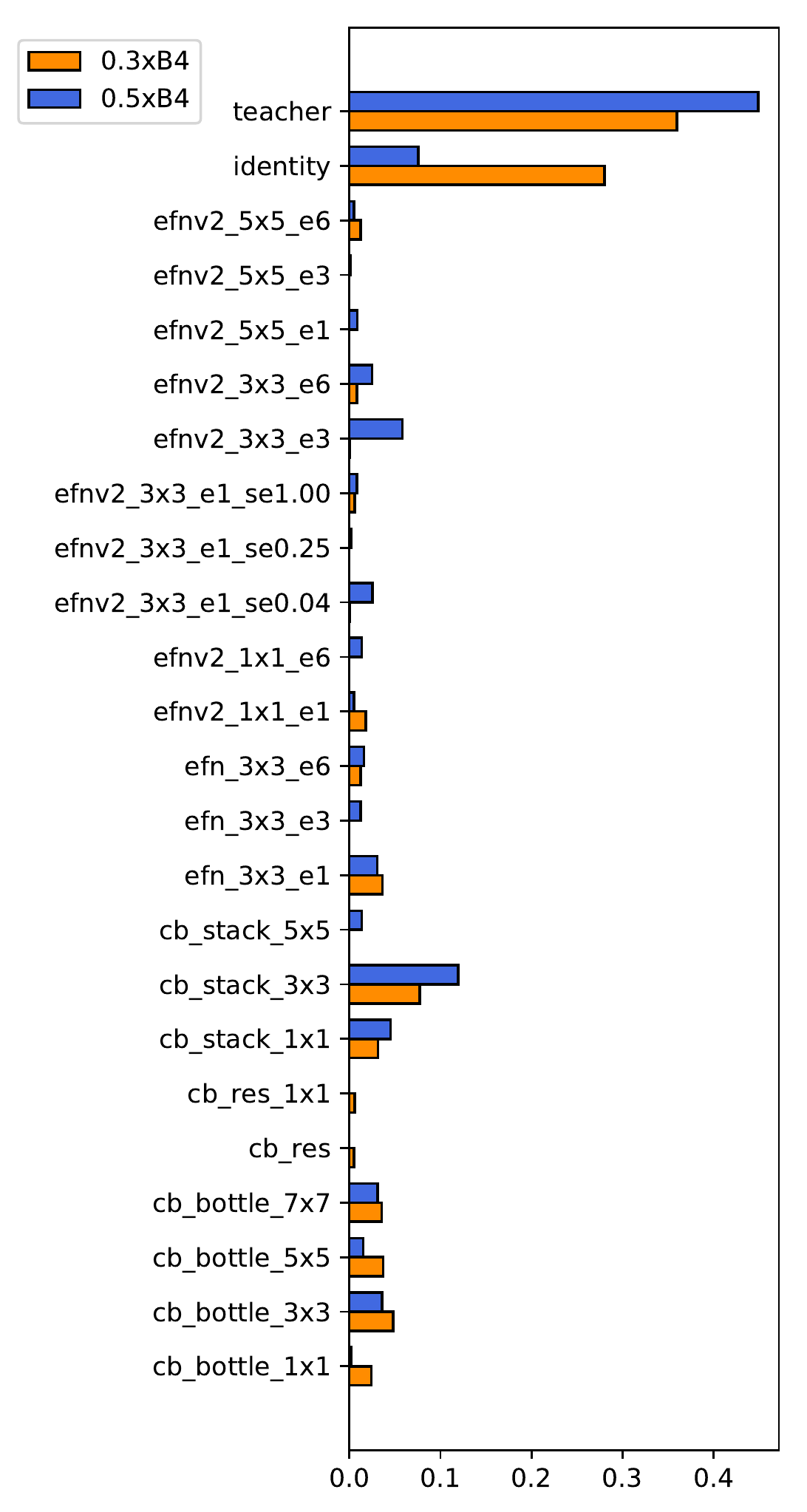}
         \caption{B4 selected ops}
         \label{fig:ops_b4}
     \end{subfigure}
     \hfill
     \begin{subfigure}[b]{0.32\textwidth}
         \centering
         \includegraphics[width=\textwidth]{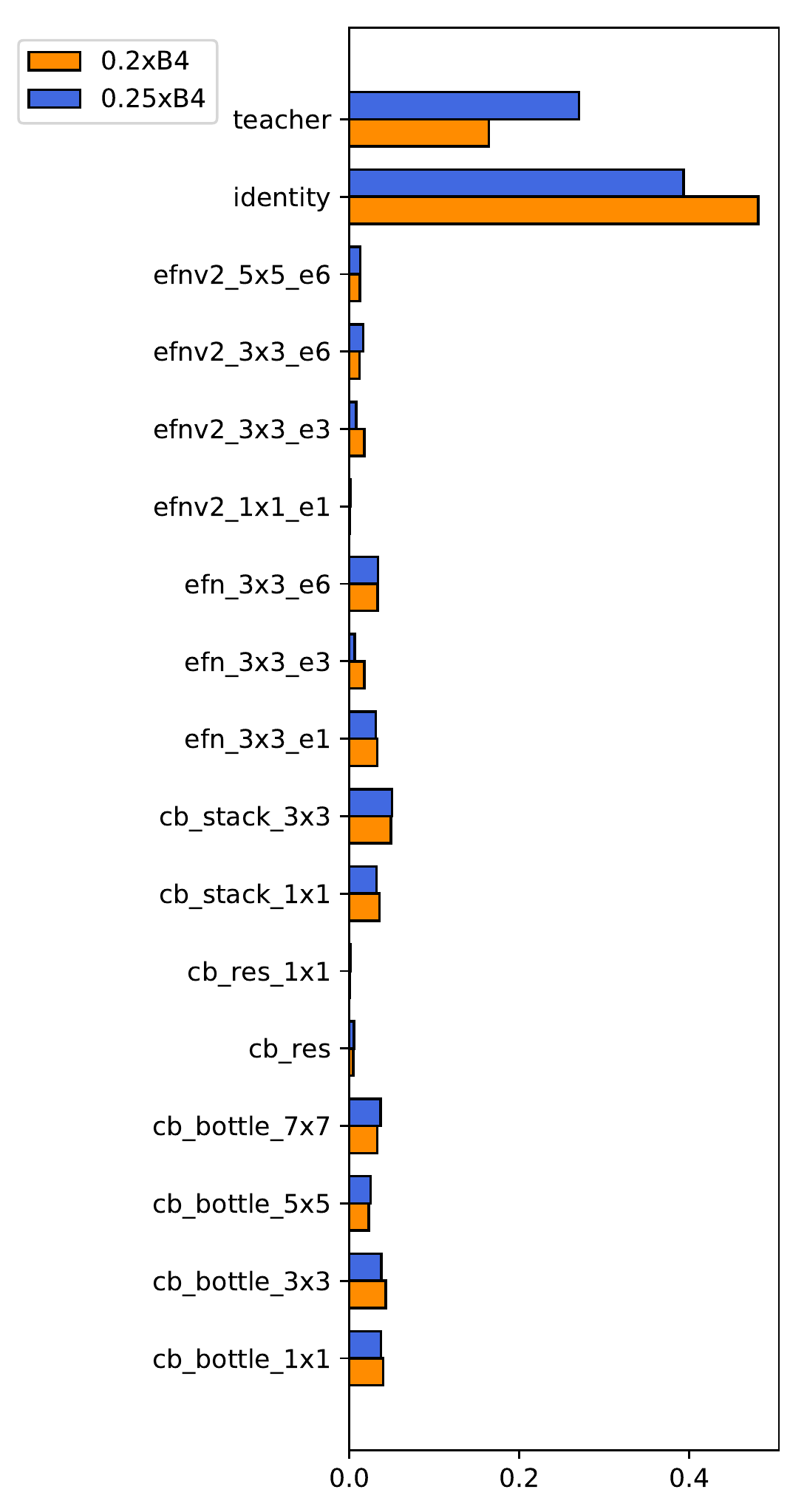}
         \caption{B4 selected ops}
         \label{fig:ops_b6}
     \end{subfigure}
     \vspace{1cm}
        \caption{The histogram of selected operations for top-100 models of EfficientNetV1 derivatives. Teacher layers are often selected especially in the deeper layers of the network as we visualize in Fig.~\ref{fig:final_arch} and Fig.~\ref{fig:final_arch2}. The identity layer is also selected often especially when the target latency is low. Interestingly, simple layers with two stacked convolution (\texttt{cb\_stack}) in the CBRCB structure (C-convolution, B-batchnorm, R-ReLU) are selected most frequently after the teacher and identity operations. Additionally we see a higher chance of selecting inverted residual blocks (\texttt{efn} and \texttt{efn2}) with no squeeze-and-excitation operations \texttt{se1.00}. }
        \label{fig:operators_distribution}
\end{figure*}

\paragraph{Distribution of selected operations.} In Fig.~\ref{fig:operators_distribution}, we provide the histogram of selected operations for three EfficientNeV1t networks and different acceleration ratios. All statistics are calculated for the first $100$ architectures found via integer optimization. 
\begin{figure*}
 \centering    
     \begin{subfigure}[b]{0.15\textwidth}
         \centering
         \includegraphics[height=.7\textheight]{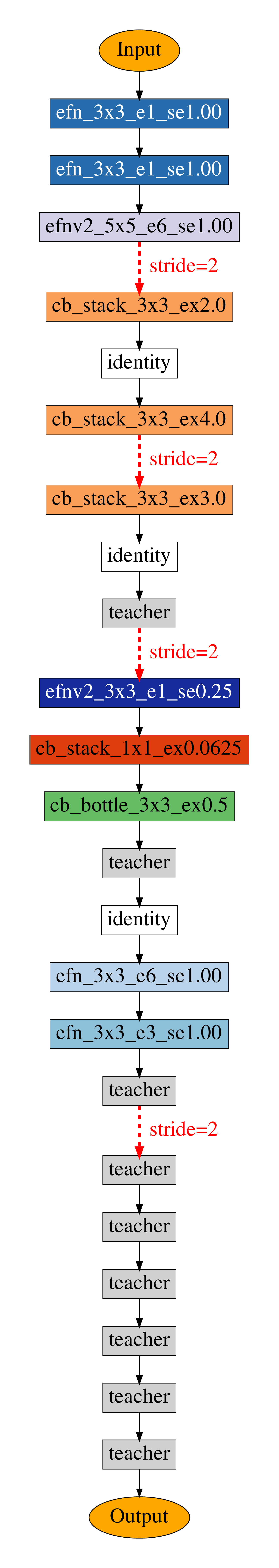}
         \caption{0.45xB2}
     \end{subfigure}
     \begin{subfigure}[b]{0.15\textwidth}
         \centering
         \includegraphics[height=.7\textheight]{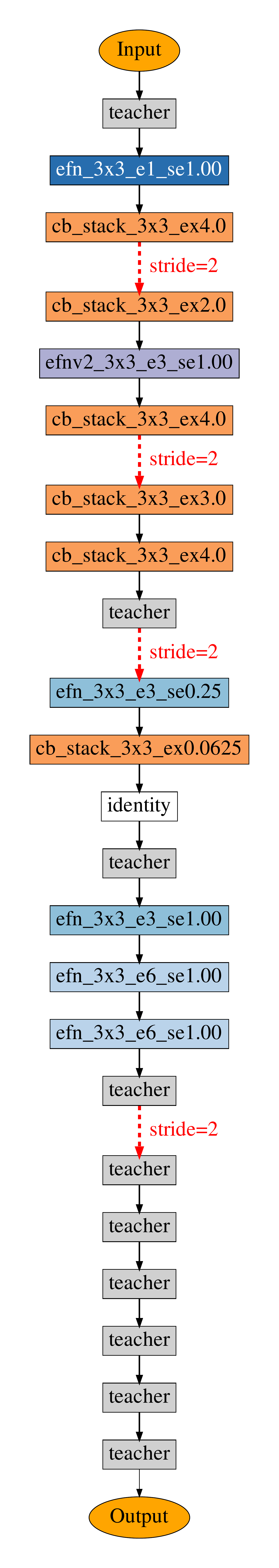}
         \caption{0.55xB2}
     \end{subfigure}
     \begin{subfigure}[b]{0.15\textwidth}
         \centering
         \includegraphics[height=.9\textheight]{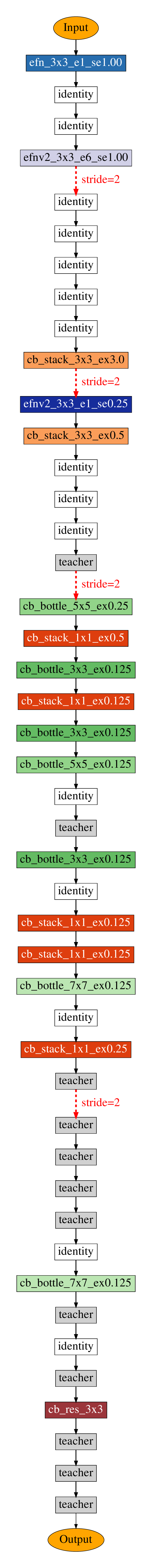}
         \caption{0.25xB6}
     \end{subfigure}
     \begin{subfigure}[b]{0.15\textwidth}
         \centering
         \includegraphics[height=.9\textheight]{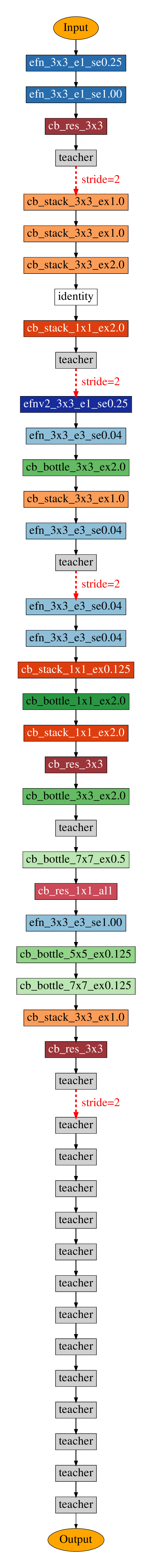}
         \caption{0.5xB6}
     \end{subfigure}
     \vspace{1cm}
        \caption{Final architectures selected by LANA  as EfficientNet-B2/B6 derivatives.}
        \label{fig:final_arch}
\end{figure*}

\begin{figure*}
\centering
     \begin{subfigure}[b]{0.15\textwidth}
         \centering
         \includegraphics[height=.9\textheight]{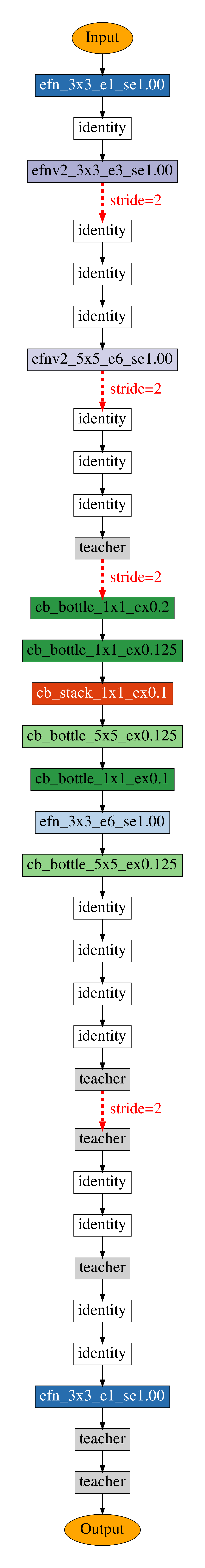}
         \caption{0.2xV1-B4}
     \end{subfigure}
     \begin{subfigure}[b]{0.15\textwidth}
         \centering
         \includegraphics[height=.9\textheight]{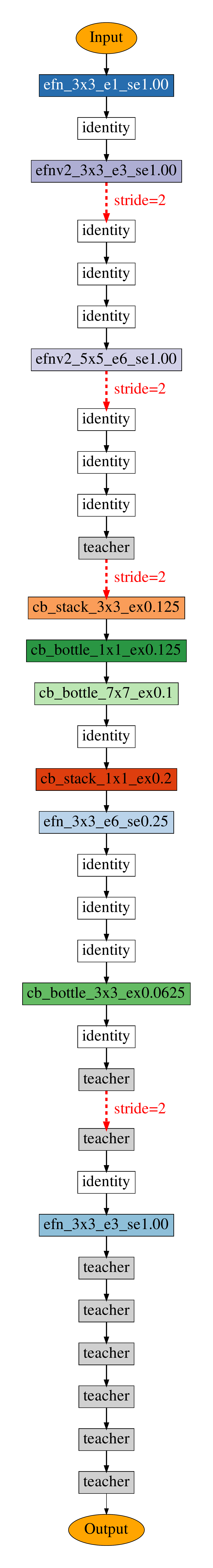}
         \caption{0.25xV1-B4}
     \end{subfigure}
          \begin{subfigure}[b]{0.15\textwidth}
         \centering
         \includegraphics[height=.9\textheight]{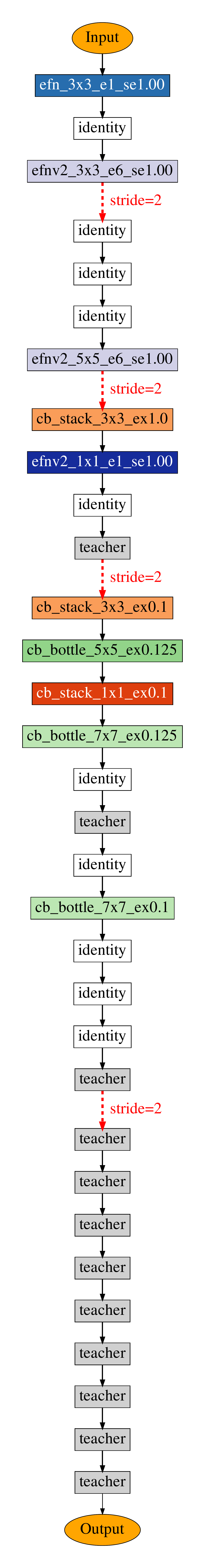}
         \caption{0.3xV1-B4}
     \end{subfigure}
     \begin{subfigure}[b]{0.15\textwidth}
         \centering
         \includegraphics[height=.9\textheight]{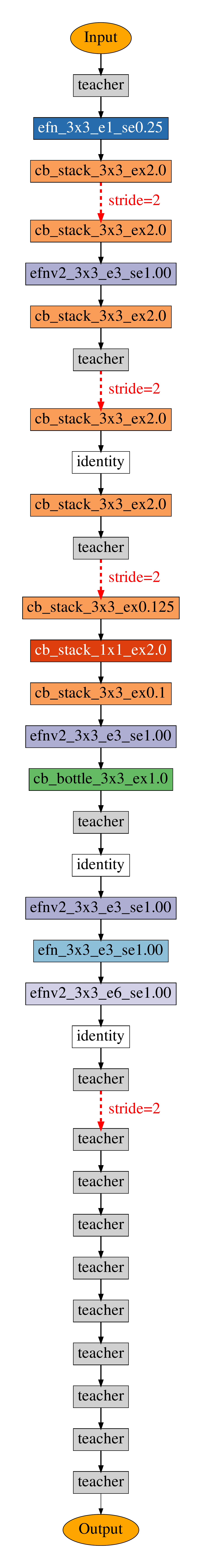}
         \caption{0.5xV1-B4}
     \end{subfigure}
     \begin{subfigure}[b]{0.15\textwidth}
         \centering
         \includegraphics[height=.9\textheight]{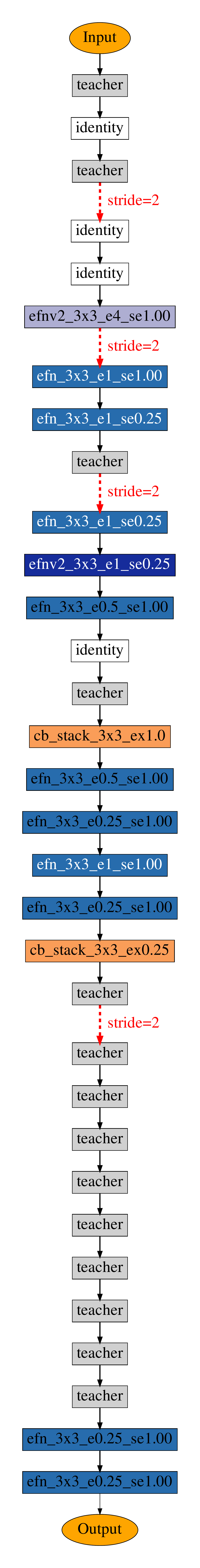}
         \caption{0.5xV2-B3}
     \end{subfigure}     
     \vspace{1cm}
        \caption{Final architectures selected by LANA  as EfficientNetV1-B4/V2-B3 derivatives.}
        \label{fig:final_arch2}
\end{figure*}

\begin{figure*}
\centering
\includegraphics[width=.9\textwidth]{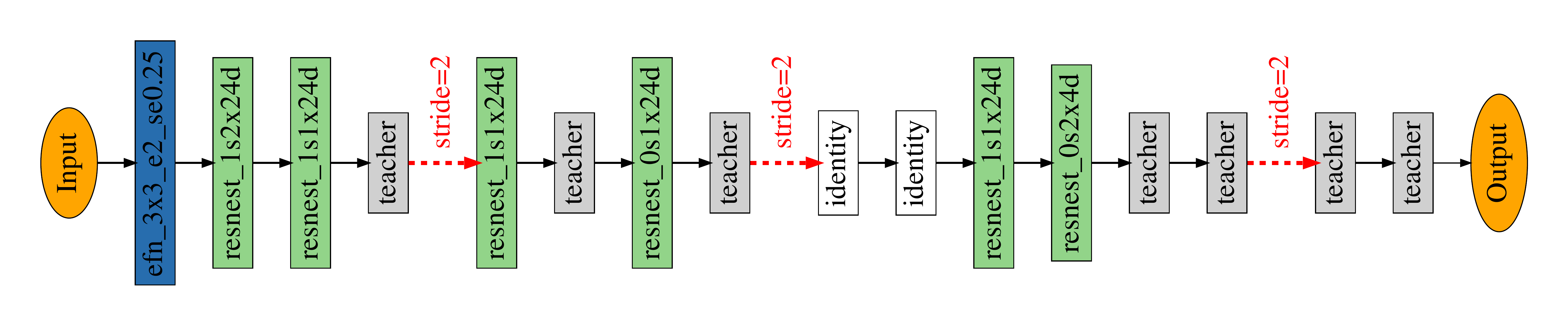}

\caption{Final architectures selected by LANA  as 0.7xResNeST50d\_1s4x24d}
\label{fig:final_arch3}
\end{figure*}

\paragraph{Final architectures.}
Fig.~\ref{fig:final_arch} and Fig.~\ref{fig:final_arch2} visualize the final architectures found by LANA. We observe that teacher ops usually appear towards the end of the networks. Identity connections appear in the first few resolution blocks where the latency is the highest to speed up inference, for example 0.2$\times$B4 has 2, 1, 1 in the first 3 resolution blocks from original 3, 4, 4. 

\subsection{Additional implementation details}
We next provide details on chosen batch size defined as \texttt{bs} and learning rate \texttt{lr}, joint with other details required to replicate results in the paper. 

\paragraph{Pretraining implementation.} Pretraining stage was implemented to distill a single operator over all layers in parallel on 4xV100 NVIDIA GPU with 32GB. For EfficientNet-B2 we set \texttt{lr}=$0.008$ with \texttt{bs}=$128$, for EfficientNet-B4 \texttt{lr}=$0.0005$ with \texttt{bs}=$40$, and EfficientNet-B6 \texttt{lr}=$0.0012$ with \texttt{bs}=$12$. We set $\gamma_\textrm{MSE}=0.001$. We run optimization with an SGD optimizer with no weight decay for   1 epoch only. 

\paragraph{Finetuning implementation.} Final model finetuning runs for 100 epochs. We set \texttt{bs}=$128$ and \texttt{lr}=$0.02$ for EfficientNet-B2 trained on 2x8 V100 NVIDIA GPU; for EfficientNet-B4 derivatives we set \texttt{bs}=$128$ and \texttt{lr}=$0.04$, for EfficientNet-B6 \texttt{bs}=$48$ and \texttt{lr}=$0.08$ on 4x8 V100 NVIDIA GPU. Learning rate was set to be $0.02$. We set $\gamma_{\text{CE}}$ and $\gamma_{\text{KL}}$ to 1.

\paragraph{Latency look up table creations.} We measure the latency on V100 NVIDIA GPU with TensorRT in FP16 mode for batch size of 128 images. For Xeon CPU latency we use a batch size of 1. Input and output stems are not included in latency LUT. This results in a small discrepancy between theoretical and real speed. As a result, we use latency LUT for operator evaluation, and report the final real latency for the unveiled final models.

\subsection{Candidate pretraining insights}
\label{sec:pretraining_app}
\begin{figure*}
    \centering
    \includegraphics[width=0.9\textwidth,trim=.25cm .25cm .25cm .25cm,clip=true]{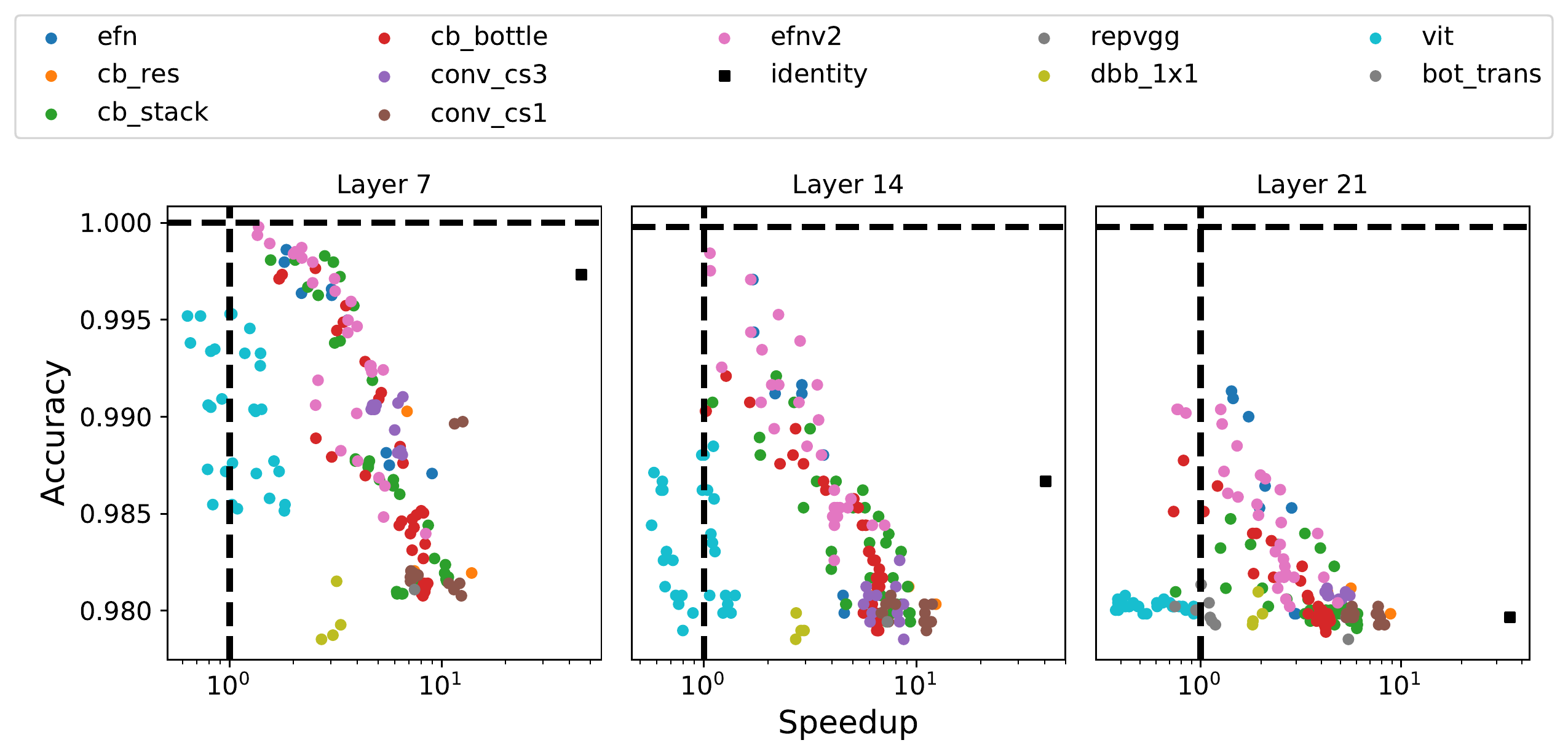}
    \vspace{0.25cm}
    \caption{Result of the pretraining stage for EfficientNetB2, showing three layers equally spaced throughout the network: 7, 14 and 21. Speedup is measured as the ratio between the latency of the teacher and the latency of the student operation (higher is better). We measure latency using Pytorch FP16. Accuracy is the ratio of the operation's accuracy and the teacher's (higher is better). The dashed black lines correspond to the teacher.} 
    \label{fig:phase2}
\end{figure*}

Latency-accuracy tradeoff for different operations after pretraining is shown in the Figure~\ref{fig:phase2}. Observations from these plots are discussed in Seection~\ref{sec:pretraining_insights}.

\textbf{The choice of pretraining loss.} To motivate our choice of MSE for pretraining, we investigate the distribution of activations at the output of residual blocks. We observe that for all blocks, activations follow a Gaussian-like distribution. Shapiro-Wilk test for normality averaged over all layers is 0.99 for EfficientNetV1-B2, and 0.988 for EfficientNetV2-B3. Given this observation, MSE error seems a reasonable loss function to minimize.

\subsection{Detailed comparison to prior work}
For comparison to prior work we look into latest models from the out-of-the-box \emph{timm package}~\cite{rw2019timm} with Apache-2.0 License. We include detailed individual method names and references as follows:

\parskip=2pt
\begin{itemize}[topsep=1pt,itemsep=1pt,partopsep=0pt,parsep=0pt,leftmargin=0pt]
    \item[] \texttt{efficientnet}: Efficientnet~\cite{tan2019efficientnet}.
    \item[] \texttt{cait}: Class-attention in image transformers~\cite{touvron2021going}.
    \item[] \texttt{cspnet}: Cross-stage partial network~\cite{wang2020cspnet}.
    \item[] \texttt{deit}: (Data-efficient) vision transformer~\cite{touvron2020deit}.
    \item[] \texttt{dla}: Deep layer aggregation~\cite{yu2018deep}.
    \item[] \texttt{dpn}: Dual-path network~\cite{chen2017dual}.
    \item[] \texttt{ecanet}: Efficient channel attention network~\cite{wang2020eca}.
    \item[] \texttt{hrnet}: High-resolution network~\cite{wang2020deep}.
    \item[] \texttt{inception}: Inception V3~\cite{szegedy2016rethinking} and V4~\cite{szegedy2017inception}.
    \item[] \texttt{mixnet}: MixConv-backed network~\cite{tan2019mixconv}.
    \item[] \texttt{ofa}: Once-for-all network~\cite{cai2020once}.
    \item[] \texttt{pit}: Pooling Vision Transforms \cite{heo2021rethinking}.
    \item[] \texttt{regnetX}: Regnet network \cite{radosavovic2020designing}, accuracy is taken from the original paper.
    \item[] \texttt{regnetY}: Regnet network \cite{radosavovic2020designing} with squeeze-and-excitation operations, accuracy is taken from the original paper.
    \item[] \texttt{repvgg}: RepVGG \cite{ding2021repvgg}.
    \item[] \texttt{resnest101\_e}: Resnest101 (with bag of tricks) \cite{he2018bag}.
    \item[] \texttt{resnest50\_d}: Resnest50 (with bag of tricks) \cite{he2018bag}. 
    \item[] \texttt{resnet50\_d}: Resnet50 (with bag of tricks) \cite{he2018bag}.
    \item[] \texttt{resnetrs10\_1}: Resnet rescaled~\cite{bello2021revisiting}.
    \item[] \texttt{resnetrs15\_1}: Resnet rescaled~\cite{bello2021revisiting}.
    \item[] \texttt{resnetrs5\_0}: Resnet rescaled~\cite{bello2021revisiting}.
    \item[] \texttt{resnext50d\_32x4d}: Resnext network (with average pooling downsampling)~\cite{DBLP:journals/corr/XieGDTH16}.
    \item[] \texttt{seresnet5\_0}: Squeeze Excitement Resnet50~\cite{hu2019squeezeandexcitation}.
    \item[] \texttt{skresnext50\_32x4d}: Selective kernel Resnext50~\cite{li2019selective}.
    \item[] \texttt{vit-base}: Visual Transformer, base architecture.
    \item[] \texttt{vit-large\_384}: Visual Transformer, large architecture, 384 resolution~\cite{dosovitskiy2020image}.
    \item[] \texttt{wide\_resnet50\_2}: Resnet50 with \(2\times\) channel width~\cite{DBLP:journals/corr/ZagoruykoK16}.
    \item[] \texttt{xception6\_5}: Xception network (original)~\cite{chollet2017xception}.
    \item[] \texttt{xception7\_1}: Xception network aligned~\cite{chen2018encoderdecoder}.
\end{itemize}

\begin{figure*}
    \centering
    \begin{tabular}{cc}
       PyTorch FP16:  &  TensorRT:\\
        \includegraphics[width=0.5\textwidth, trim={0 0 0px 0},clip]{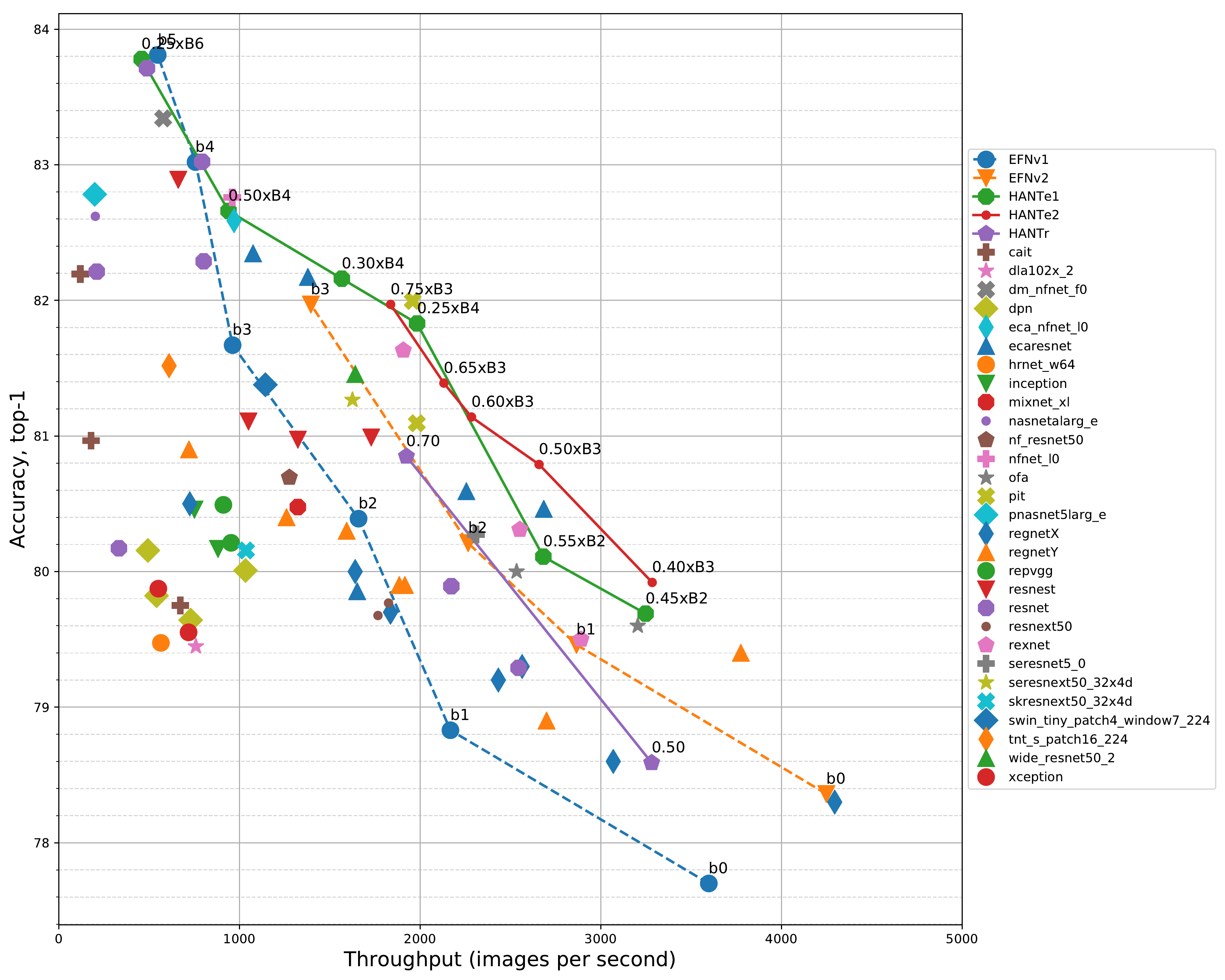} & \includegraphics[width=0.5\textwidth, trim={0 0 0px 0},clip]{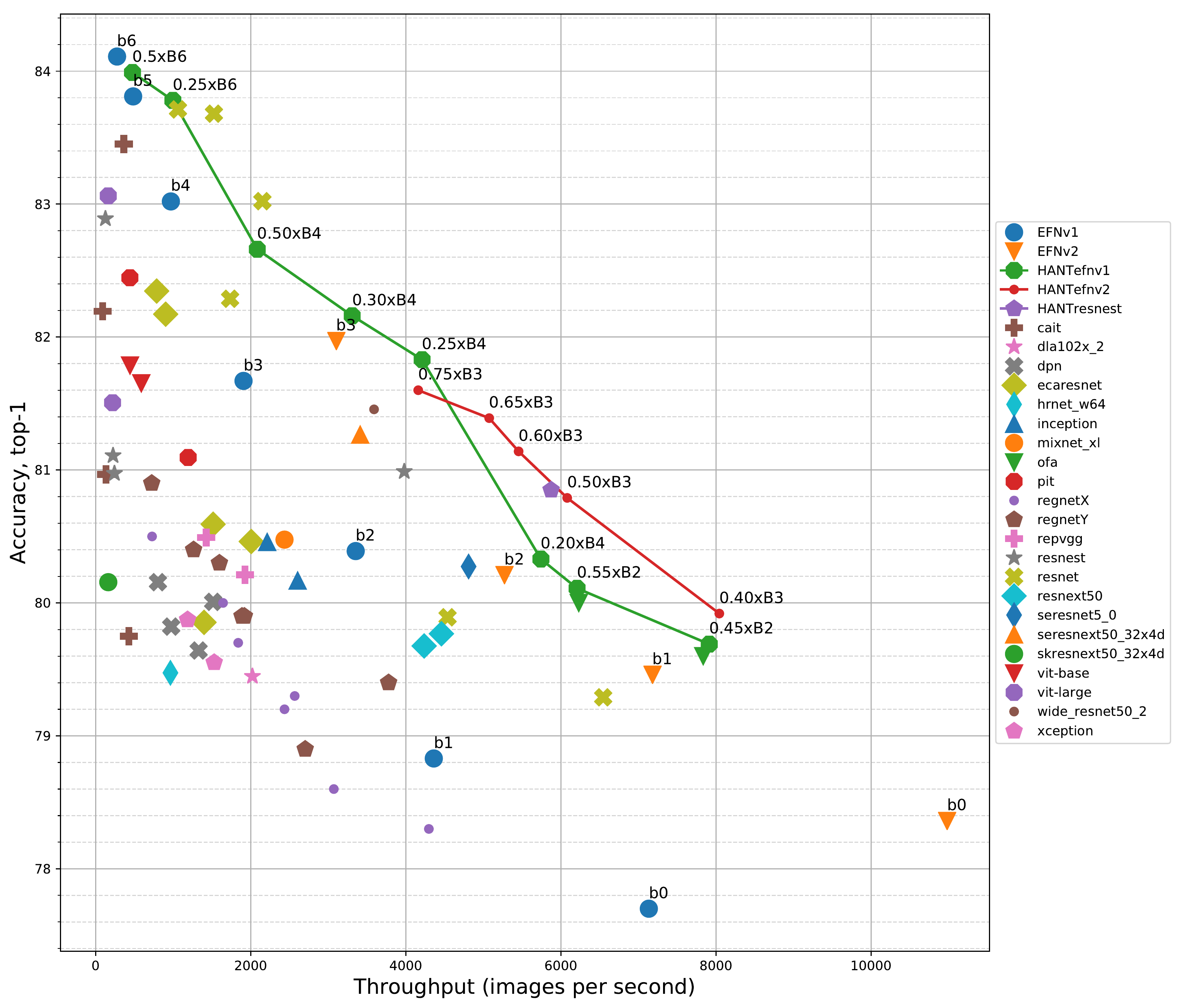}
    \end{tabular}
    \caption{Comparison with other models from \texttt{TIMM} package.}
    \label{fig:imagenet_more}
\end{figure*}

A more detailed comparison with other models is shown in the Figure~\ref{fig:imagenet_more}. We observe that models resulted from LANA acceleration are performing better than the most of other approaches. All of the models for LANA used LUTs computed with TensorRT and clearly the speed up in the TensorRT figure is larger when compared with other methods. On the same time if model latency is estimated with Pytorch, we still get top models that outperform many other models. 

\end{document}